# HAMLET: Healthcare-focused Adaptive Multilingual Learning Embedding-based Topic Modeling

**Hajar Sakai and Sarah S. Lam**
School of Systems Science and Industrial Engineering
State University of University at Binghamton
Binghamton, NY, USA
hsakai1, sarahlam@binghamton.edu

## Abstract

Traditional topic models often struggle with contextual nuances and fail to adequately handle polysemy and rare words. This limitation typically results in topics that lack coherence and quality. Large Language Models (LLMs) can mitigate this issue by generating an initial set of topics. However, these raw topics frequently lack refinement and representativeness, which leads to redundancy without lexical similarity and reduced interpretability. This paper introduces HAMLET, a graph-driven architecture for cross-lingual healthcare topic modeling that uses LLMs. The proposed approach leverages neural-enhanced semantic fusion to refine the embeddings of topics generated by the LLM. Instead of relying solely on statistical co-occurrence or human interpretation to extract topics from a document corpus, this method introduces a topic embedding refinement that uses Bidirectional Encoder Representations from Transformers (BERT) and Graph Neural Networks (GNN). After topic generation, a hybrid technique that involves BERT and Sentence-BERT (SBERT) is employed for embedding. The topic representations are further refined using a GNN, which establishes connections between documents, topics, words, similar topics, and similar words. A novel method is introduced to compute similarities. Consequently, the topic embeddings are refined, and the top k topics are extracted. Experiments were conducted using two healthcare datasets—one in English and one in French—from which six sets were derived. The results demonstrate the effectiveness of HAMLET.

## 1. Introduction

Whether it be medical literature, clinical notes, or healthcare communications, the volume of healthcare text data collected daily is growing exponentially. As a result, various opportunities have emerged, which are accompanied by significant challenges when it comes to extracting relevant insights. One key Natural Language Processing (NLP) task that enables that converts raw data into potentially actionable information is text classification. However, several real-world scenarios lack labels, which means prior topic modeling is required to identify the topics that can serve as labels (Figure 1).

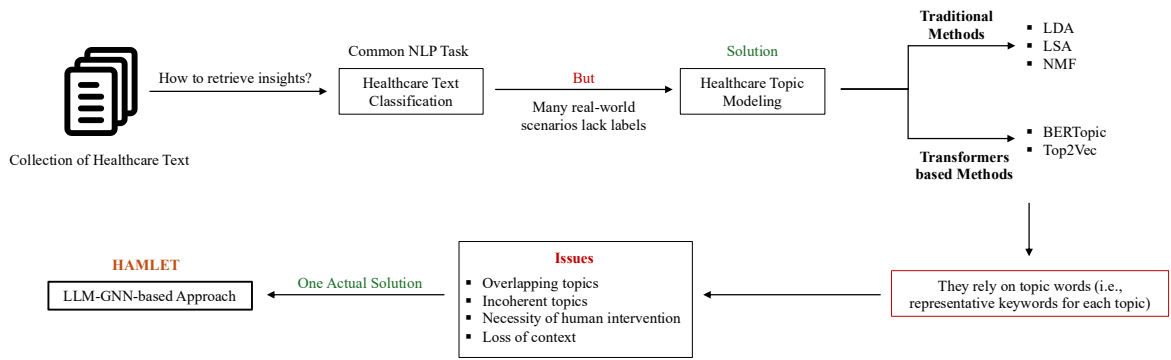

**Figure 1.** Research Problem

Multiple existing topic models, such as Latent Dirichlet Allocation (LDA) (Blei et al., 2003), are useful; however, they typically struggle to establish clear topic separation. Traditional topic models often fail to distinguish topics due to word frequency overlaps and semantic ambiguity, which result in less distinctive topics that limit downstream tasks such as text classification (Figure 1). This paper proposes a novel framework that integrates GNNs and LLMs to improve topic modeling, particularly in healthcare settings where the data is characterized by its complexity.

LLMs have a proven record of suitability for multiple healthcare-related NLP tasks (Nassiri & Akhloufi, 2024; Bedi et al., 2024; Li et al., 2024). From domain-specific fine-tuned models such as BioBERT (Lee et al., 2020) and PubMedBERT (Gu et al., 2022) to general-purpose models such as GPT-4o (OpenAI, 2024), these models have demonstrated a notable ability to understand healthcare context in textual data. While LLMs' strength lies in language understanding, primarily designed to operate on sequential data and rely on the attention mechanism to capture contextual relationships, they often lack the capability of explicitly modeling the inherent graph-like structure present in the data. GNNs, in contrast, are distinguished by their ability to deal with graph-structured data (i.e., graph networks), which captures the complex relationships between the different entities (e.g., texts and topics) (Hamilton et al., 2017). Combining the contextual understanding of an LLM and the structural learning that a GNN brings, the proposed topic model leverages the strengths of each component of this hybrid approach. The LLM (i.e., GPT-4o) is employed to directly process the raw healthcare text and generate an initial pool of unpolished and redundant topics. At the same time, the GNN operates on the hybrid graph network constructed to refine the topics' embeddings, which leads to the extraction of the top k (i.e., k = 50 or 25) relevant topics discussed in the dataset considered.

The main contributions of this research are as follows:

- **SBERT-BERT Hybridization**: Embedding text documents and topics' phrases using a hybrid approach combining BERT's (Devlin et al., 2018) contextual strength and SBERT's (Reimers & Gurevych, 2019) sentence-level semantic capabilities.
- **Semantic-Geometric Similarity (SGS) Method**: Proposing a hybrid similarity method based on Word Mover's Distance-inspired and Cosine similarities for computing similarities among topics and words.
- **Hybrid Graph Network Construction**: Creating a graph network with nodes representing SBERT-BERT-based document and topic embeddings, and BERT-based word embeddings. Establishing edges between topics, documents, words, and similar topics/words.
- **EdgeRefineGNN**: Introducing a GNN-based topics' refinement pipeline using edge-conditioned convolution to integrate different nodes and leverage edge-specific attributes.
- **Top k Topics Extraction**: Conducting clustering followed by coherence-based extraction of the top k topics.

The experiments presented in this paper utilize two source datasets. The first dataset, in English, comprises patient experience comments gathered through surveys, while the second, in French, consists of COVID-19-related French news articles. Three subsets were sampled from each source dataset, resulting in six datasets varying in the average number of tokens per text document. These two source datasets were chosen to reflect real-world healthcare scenarios where topics must first be explored, investigated, and identified before text classification, while demonstrating HAMLET's cross-lingual capabilities.

This research is structured as follows: The literature about leveraging LLMs and GNNs for healthcare topic modeling is reviewed in the "Related Literature" section. The "Data and Sampling" section describes the datasets used as well as the sampling carried out. HAMLET, the methodology developed in this research, is detailed in the "Proposed Approach" section. The results of both HAMLET and existing topic models are discussed in the "Results and Discussion" section. Lastly, "Conclusion and Future Directions" presents the conclusions and future work.

## 2. Related Literature
Large Language Models (LLMs) changed how multiple NLP tasks are handled, which include topic modeling and text classification. The abundant text data constantly generated in the healthcare industry constitutes an exciting opportunity to investigate and attempt to convert this unexplored data into insightful information that could drive future advancements. This section reviews the key literature that developed LLM-based approaches for topic modeling, which focus on healthcare. The use of GNNs for topic modeling is also examined.

## 2.1. Large Language Models for Healthcare Topic Modeling

With LLMs, new approaches for for topic modeling have come to light. This resulted in shifting from traditional statistical approaches such as LDA and Non-negative Matrix Factorization (NMF) (Lee and Seung, 1999) relying on the identification of word co-occurrence patterns and statistical distributions to semantically rich and contextually aware LLM-based approaches. In healthcare, particularly, leveraging LLMs for topic modeling has slowly started gaining researchers' interest during the last year. The literature includes multiple research papers that investigate various strategies for adapting LLMs to topic modeling. A literature summary is outlined in Table 1.

**Table 1.** Literature Review: LLMs for Healthcare Topic Modeling

| Reference | LLM Type | LLM Contribution |
|---|---|---|
| Chopra et al. (2024) | GPT-3.5-turbo | Generating one-line descriptions for each topic, which result from LDA, based on the top 30 representative words in addition to short topic names to enable better interpretability |
| Borazio et al. (2024) | GPT-4 | Converting the generated and identified Subject-Verb-Object that represents the topics into natural language descriptions and titles that would be used afterward for classification |
| Wosny and Hastings (2024) | GPT-3.5-turbo & Mixtral 7x8b | First, extracting the themes from the individual transcripts, and afterward summarizing and integrating the topics with the dataset |
| Bitaraf et al. (2024) | Mistral-7B-Instruct-v0.2 | Identifying the top 10 topics based on the titles and abstracts of research papers |
| Islam and Goldwasser (2024) | SBERT & GPT-4 | Using SBERT for text embeddings that are clustered and GPT-4 for clusters' coherence check, summarization, and naming, resulting in new topics discovery |
| Koloski et al. (2024) | BERTopic & LLaMa2 | Clustering and keywords extraction using BERTopic. Subsequently, keywords transform into meaningful semantic labels using LLaMa2 |
| Campos et al. (2024) | GPT-3.5 | Identifying single and distinct topics from each text using prompt engineering |
| Mu et al. (2024) | GPT-3.5 & LLaMA-2-7B | Generating topics (i.e., three topics) by text documents given seeds, and subsequently summarizing them into a final list of topics |
| Rijcken et al. (2023) | GPT-3 | Outputting labels and justification based on the bag-of-words that represents the topics generated using Latent Semantic Analysis (LSA) |

Table 1 summarizes the existing literature where LLMs are used for healthcare topic modeling. Their applications vary between refining or interpreting generated topics and direct extraction of the topics, in which the dominant type of LLM used is the GPT family. However, the literature scoping reveals a significant gap in the literature, especially in niche areas such as healthcare.

## 2.2. Graph Neural Networks for Topic Modeling

Due to the limited research on GNNs for healthcare topic modeling, the literature review was expanded to include applications across other domains. Table 2 presents a summary.

Table 2. Literature Review: GNNs for Topic Modeling

| Reference | GNN Type | GNN Contribution |
|---|---|---|
| Adhya and Debarshi (2024) | Graph Isomorphism Network (GIN) | Capturing word relationships in text documents by modeling them as graphs (where words are nodes and semantic similarities are edges) for representation learning, before aggregating the output with TF-IDF representation and feeding the combined embedding to a Variational Autoencoder (VAE) |
| Sun et al. (2023) | Graph Convolution Network (GCN) | Modeling multi-role interactions in conversations by capturing intra-speaker and inter-speaker dependencies within utterances, resulting in integrating the hierarchical structure to discover coherent and interpretable topics |
| Zhou et al. (2020) | GCN | Capturing word-word and document-document relationships by representing both documents and words as nodes, which are connected based on document-word co-occurrences, and enriching document representations through information aggregation from neighboring nodes using graph convolution |
| Yang et al. (2020) | Graph Attention Network (GAT) | Using bipartite graph structure with graph attention mechanism for word embedding and document structure while integrating word similarity and word co-occurrence structure using graph convolution operations |
| Zhu et al. (2018) | GCN | Representing biterms (i.e., word pairs) as graphs with words as nodes and their co-occurrence counts as weighted edges. Resulting in capturing transitive relationships between words and therefore extracting coherent topics |

As shown in Table 2, the literature reveals that GNNs have emerged as a suitable technique for topic modeling by capturing different types of complex relationships within textual data. Using different types of GNNs (i.e., isomorphism, convolution, attention) demonstrates their ability to preserve the structural information contained in text data that traditional topic modeling methods often miss.

### 2.3. LLMs and GNNs for Topic Modeling

While some research has examined the combination of LLMs and GNNs for topic modeling, their application to healthcare text datasets remains unexplored. In the few studies that incorporated language models and graph-based approaches into topic modeling frameworks, researchers primarily used graph embedding algorithms such as node2vec (Loureiro et al., 2023) or graph-based clustering (Altuncu et al., 2021) rather than GNNs. Furthermore, the contribution of LLMs in these studies was limited to generating text embeddings, such as those from XLM-RoBERTa (Mohawesh et al., 2023), SBERT (Loureiro et al., 2023; Altuncu et al., 2021), and bert-as-a-service (Altuncu et al., 2021). The only study that leveraged both an LLM and a GNN—specifically, a Graph Attention Network—was Mohawesh et al. (2023), but their focus was binary classification, in which topics serve only to refine document embeddings for news articles.

### 3. Data and Sampling

The proposed approach, HAMLET, conducts topic modeling on healthcare datasets in English and French. In real-world healthcare settings, labels are usually missing, necessitating topic modeling. Three datasets were sampled from each source dataset for the experiments and performance assessment.

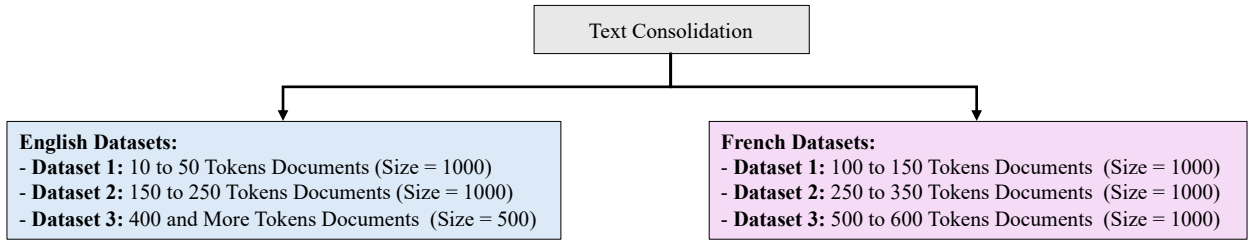

**Figure 2.** Datasets Sampled Details

The English source dataset consists of patient feedback about general practitioners and hospitals in the United Kingdom (UK), which are initially separated. Before any analysis, both comment datasets were consolidated. Each feedback comprised terms such as "Comment Title", "Liked", "Disliked", and "Advice"; the preprocessing included removing these phrases and dropping cases where the responses were "Nothing" or "N/A". By examining the number of tokens (i.e., word count) distribution of the preprocessed text documents (i.e., Figure 3), three samples of different token ranges were retrieved. The goal is to assess how HAMLET performs as the average document length varies.

The French source dataset consists of COVID-19-related French-speaking online newspapers from various online newspapers. The data includes the title, description, and text for each article. These different components were consolidated, which results in the analyzed textual documents. Like the English dataset, three datasets were sampled based on the number of tokens. Figure 4 shows the distribution of document word counts.

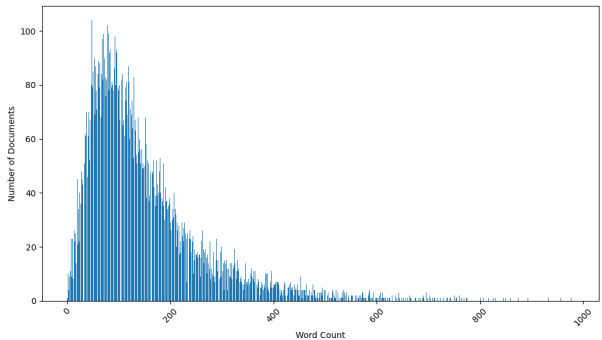
**Figure 3.** [English] Distribution of Document Word Counts

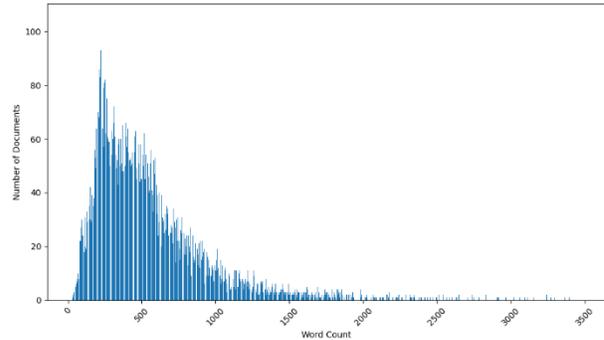
**Figure 4.** [French] Distribution of Document Word Counts

The source datasets differ in size—the English dataset contains 15,380 documents while the French dataset contains 40,434—and has different token distributions, which explains the ranges and sizes of the sample datasets shown in Figure 2. The sets involved in this research enable not only evaluating the effectiveness of the proposed approach, HAMLET, but also assessing its performance across two different languages, English and French, and across text documents varying from short to medium to long.

## 4. Proposed Approach

This section introduces a framework for analyzing healthcare textual data in English and French through topic modeling. Figure 5 outlines the steps contributing to the proposed LLM-Graph-based topic modeling approach, HAMLET. The study uses six healthcare datasets to evaluate three key aspects: the approach's reliability, the effect of text length (measured in tokens), and performance consistency across two languages.

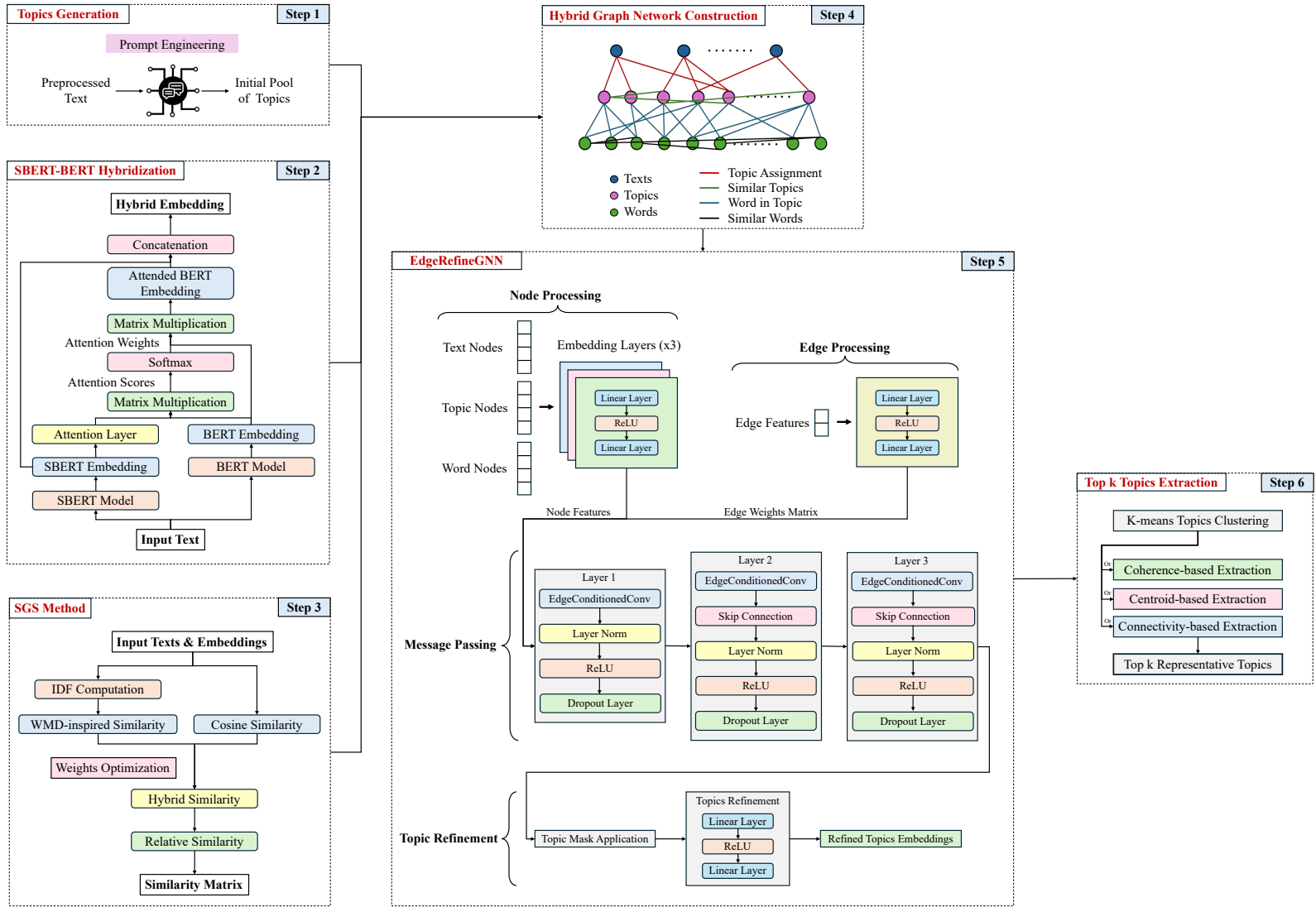

**Figure 0.** Proposed HAMLET Approach

## 4.1. Topics Generation

HAMLET involves leveraging an LLM to conduct topic generation by document. The LLM used is GPT-4o. As a general-purpose language model, GPT-4o's advanced language understanding, contextual awareness, and accessibility through API requests made it a suitable candidate for this case study. The topic generation was conducted via prompt engineering (Figures 6 and 7).

---

input = f""""{text}"""
prompt = f""""You are a healthcare expert tasked with analyzing patient feedback. Your task is to:
  1. **Sentiment Analysis**: Assign a relevant sentiment to each comment in the list of patient feedback provided. The sentiment can be:
     - Positive
     - Negative
     - Neutral
     - Mixed
  2. **Topic Identification**: Identify and list the specific and meaningful topics mentioned in each patient feedback comment. The topics should reflect the patient's complaints and can have only a negative connotation.
  Please provide the results in JSON format with the following keys:
     - Comment
     - Sentiment
     - Topics
  Here are the comments to analyze:
  ```{input}```
  """

---

**Figure 6.** English Dataset Topics Generation Prompt

---

input = f""""{text}"""
prompt = f""""Vous êtes un expert en santé chargé d'analyser des données textuelles relatives au domaine médical et à l'impact du COVID-19. Les données textuelles sont des articles d'actualité. Votre tâche consiste à :
  1. **Analyse des sentiments**: Attribuer un sentiment pertinent à chaque texte dans la liste des données textuelles fournies. Le sentiment peut être :
     - Positif
     - Négatif
     - Neutre
     - Mitigé
  2. **Identification des sujets**: Examiner et identifier les thèmes ou sujets spécifiques mentionnés dans chaque texte, tels que la violence sur les réseaux sociaux, la désinformation médicale, les menaces envers les professionnels de la santé, l'éthique médicale, l'impact sur l'économie, et les faits liés au sport.
  Veuillez structurer les résultats au format JSON avec les clés suivantes :
  - ID
  - Sentiment
  - Sujets
  Chaque texte peut se voir attribuer un ou plusieurs thèmes/sujets. Les thèmes/sujets peuvent consister sur un mot ou des phrases très courtes.
  Les thèmes/sujets doivent refléter des labels qui pourront par la suite être utilisés pour classifier ces données textes et ne doivent pas être des mots clés aléatoires.
  Voici les textes à analyser :
  ```{input}```
  """

---

**Figure 7.** French Dataset Topics Generation Prompt

The English and French prompts are structured to include contextualization, the tasks that the LLM should conduct, the instructions, and the expected output format. In both cases, the tasks consist of assigning the relevant sentiment and identifying each text document's topics for subsequent classification. For the patients' comments, the focus was on topics with negative connotations, assuming that the end goal is to pinpoint improvement opportunities within the corresponding hospital. For the COVID-related news, the only condition imposed is that the topics should be meaningful short phrases rather than random keywords. Multiple topics can be assigned to each document; however, a document may have no topics assigned if none are found to be relevant.

As a result, an initial pool of topics is generated. Each initial pool is large in size and therefore not suitable for downstream tasks such as text classification. Additionally, manually identifying the top k topics will lead to biases, errors, and will be an extensive task. This means that proper topic extraction is required. However, before proceeding, a GNN will be leveraged to refine the topic embeddings taking into account edge-specific attributes (e.g., relationships and similarities). Prior to that, embedding vectors for text documents and topics' phrases are obtained using a proposed hybridization of BERT and SBERT.

**Table 3.** Topics Generation-related Counts

|         | Dataset   | Dataset Size | Number of Text with Topics | Initial Pool of Topics' Size |
|---------|-----------|--------------|----------------------------|------------------------------|
| English | Dataset 1 | 1000         | 419                        | 668                          |
|         | Dataset 2 | 1000         | 127                        | 257                          |
|         | Dataset 3 | 500          | 23                         | 59                           |
| French  | Dataset 1 | 1000         | 1000                       | 1219                         |
|         | Dataset 2 | 1000         | 999                        | 1158                         |
|         | Dataset 3 | 1000         | 1000                       | 1267                         |

## 4.2. SBERT-BERT Hybridization

The involvement of a GNN requires adequately converting all types of texts in this topic modeling into representative vectors. In this case, there are three types of texts: the text documents representing the dataset, the topics generated in Step 1, and the words contained in each topic. For word embeddings, BERT is used. However, SBERT-BERT Hybridization, as shown in Algorithm 1, is employed for the text documents and topic phrases.

SBERT-BERT hybridization combines SBERT and BERT embeddings through an attention mechanism. The input text is processed through both models to obtain the initial respective sentence-level and token-level embeddings. Subsequently, an attention layer is introduced to calculate the weights and identify which tokens in BERT's embedding are most relevant to the overall sentence meaning based on SBERT's embedding. The resulting weights are afterward applied to BERT's embedding, which leads to the calculation of the attended BERT embedding. The hybrid embedding is the concatenation of the SBERT embedding and attended BERT embedding. This improved version of embeddings combines SBERT's sentence-level understanding with BERT's contextual word representation.

---

**Algorithm 1: Pseudocode of the SBERT-BERT Hybridization**

**Input:** Text
**Output:** Hybrid embedding vector
**Initialization:**
  **a. Load pre-trained SBERT model with 'bert-base-nli-mean-tokens'**
  **b. Load pre-trained BERT model with 'bert-base-uncased'**
  c. Create attention (linear) layer matching SBERT dimension to BERT hidden size
1. Generate SBERT embedding: sbert_embedding = SBERT(text)
2. Generate BERT embedding:
  a. tokenize_input = tokenize(text)
  b. bert_output = BERT(tokenize_input)
  c. bert_embedding = extract_last_hidden_state(bert_output)
3. Compute attention weights:
  a. attention_scores = matmul(attention_layer(sbert_embedding), transpose(bert_embedding))
  b. attention_weights = softmax(attention_scores)
4. Apply attention weights to BERT embedding: attended_bert = matmul(attention_weights, bert_embedding)
5. Create hybrid embedding: hybrid_embedding = concatenate(sbert_embedding, attended_bert)
6. Return the hybrid embedding

---

## 4.3. Semantic-Geometric Similarity (SGS) Method

One type of the edges that forms the constructed graph network reflects the similarities between topics. To compute these similarities, this section proposes the Semantic-Geometric Similarity (SGS), a hybrid method which integrates

Word Mover's Distance (WMD)-inspired similarity, cosine similarity, and Inverse Document Frequency (IDF) weighting. By leveraging these components, SGS seeks to capture the semantic and contextual relationships among compared topics. This method was first proposed to calculate topics' similarities but was subsequently employed for words' similarities as well.

### 4.3.1. IDF Computation

The first key component of SGS is the IDF. Each word is calculated to quantify its importance in the corpus (i.e., set of topics). For a word w in a set of topics T, the IDF is computed as follows:

$$\text{IDF}(w) = \log\left(\frac{N}{\text{count}(w)}\right)$$

where N is the total number of topics and count(w) the number of topics containing w.

The calculation of the IDF across all topics will contribute to the identification of words that should be assigned more weight when it comes to determining similarities.

### 4.3.2. Word Mover Distance-inspired Similarity

The Word Mover Distance-inspired similarity is a critical component of the SGS method. This component of the proposed method is an adaptation of the traditional Word Mover's Distance (WMD) algorithm (Kusner et al., 2015). It operates through multiple steps:

- **Vocabulary Filtering:** This initial filtering is conducted to ensure that only the words with corresponding available embeddings are considered, which therefore reduces noise and computational inefficiency.

- **IDF-Weighted Word Embeddings:** The previously calculated IDF scores are used to weight the original word embeddings using a tunable parameter weight$_{idf}$. The word embeddings are first generated using BERT. As a result, the weighting will be to distinguish words that appear in fewer topics and therefore effectively capture the relative significance of each token.

$$\text{emb}_{1i} = \text{word\_vector}[\text{word}_{1i}] \times (1 + \text{weight}_{idf} \times \text{idf}[\text{word}_{1i}])$$
$$\text{emb}_{2j} = \text{word\_vector}[\text{word}_{2j}] \times (1 + \text{weight}_{idf} \times \text{idf}[\text{word}_{2j}])$$

where 1 and 2 correspond to the topics to compare, and i is the position of each word in the topic phrase.

- **Distance Matrix Calculation:** The cosine distance matrix between all pairs of weighted word vectors from both topics is computed, creating a comprehensive comparison of semantic relationships.

$$D[i, j] = \text{cosine\_distance}(\text{emb}_{1i}, \text{emb}_{2j})$$

- **Assignment Problem Solution:** The Hungarian algorithm is used to find the optimal one-to-one alignment between words from both topics that minimizes the total semantic distance. For this purpose, linear_sum_assignment from the Python package SciPy is leveraged.

$$\text{total\_dist} = \sum D[\text{row}[k], \text{col}[k]] \text{ for k in range}(\min(|\text{split}(\text{topic}_1)|, |\text{split}(\text{topic}_2)|))$$

where row and col are the lists of optimal assignment indices returned by the Hungarian algorithm.

- **Similarity Normalization:** The calculated distance is converted to a similarity score between 0 and 1. Additionally, a normalization by the maximum number of words in either topic to account for different topic lengths is conducted:

$$\text{Similarity}_{wmd} = 1 - \frac{\text{total\_dist}}{2 \times \max(|\text{split}(\text{topic}_1)|, |\text{split}(\text{topic}_2)|)}$$

This WMD-inspired similarity is characterized by soft IDF weighting, where a soft weighting scheme via the tunable weight$_{idf}$ parameter allows for more flexible control over how much word distinctiveness influences similarity. This would not have been possible using hard IDF weights. Moreover, using a cosine distance metric instead of Euclidean distance, for instance, helps capture semantic differences while being less sensitive to the word vectors' respective magnitudes. Furthermore, the normalization by twice the maximum topic length permits the proposed method to work well even when comparing topics with significantly different lengths. As a result, this component constitutes a critical foundation of the overall SGS. It enables a fine-grained word-level comparison that will complement the document-level (i.e., topic-level) comparisons provided by the cosine similarity.

### 4.3.3. Hybrid Similarity

The core of the SGS method is the hybrid similarity calculation, in which two complementary similarity metrics are combined in a unified fashion. In addition to the fine-grained word-level matching WMD-inspired similarity, the cosine similarity, which represents a holistic document-level comparison, is computed between topic pairs.

$$\text{Similarity}_{cosine} = 1 - \text{cosine}(\text{topic}_{emb_i}, \text{topic}_{emb_j})$$

A balanced hybrid measure is afterward calculated using a weighted average of the two resulting similarity matrices. The weight$_{wmd}$ controls the balance and is also tunable.

$$\text{Similarity}_{hybrid} = \text{weight}_{wmd} \times \text{Similarity}_{wmd} + (1 - \text{weight}_{wmd}) \times \text{Similarity}_{cosine}$$

The goal is to leverage both WMD-inspired similarity, which captures the fine-grained semantic relationships that might be missed by cosine similarity alongside cosine similarity, which provides a holistic comparison that might overlook specific word alignments.

### 4.3.4. Relative Similarity

To provide a context-aware normalization of the similarity scores computed, a relative similarity is calculated. This consists of first centering them by subtracting the mean to make the values near the average similarity close to 0. The centered values are scaled by dividing by half the standard deviation. Using half the standard deviation rather than the full standard deviation amplifies the differences between similarity scores, because it sharpens the distinctions between them, which makes it easier to differentiate between close and distant pairs of topics. Lastly, the centered and scaled values are passed through a sigmoid function, which maps all the similarity scores to the range [0,1], and which reduces, as a result, the impact of the outliers.

$$\text{Similarity}_{hyb\_rlv} = \frac{1}{1 + e^{-\frac{\text{Similarity}_{hybrid} - \mu}{\sigma/2}}}$$

where, $\mu$ is the mean similarity and $\sigma$ is the standard deviation of similarities.

The relative similarity transformation provides multiple advantages to the resulting scores, which include context-awareness (i.e., adaptation to the dataset's characteristics), enhanced contrast (i.e., using halved standard deviations), outlier robustness (i.e., the sigmoid function compression), and range normalization.

### 4.3.5. Weights Optimization

As seen throughout the development of the SGS method, there are two weights, weight$_{idf}$ and weight$_{wmd}$, that need to be tuned. For this, an optimization process based on grid search is followed. The method starts with creating a linear space of possible weight$_{wmd}$ from 0.1 to 0.9, in which there are nine evenly spaced values. Additionally, with a constraint that the two weights should be complementary (i.e., approximately sum to 1), the goal is to ensure a balanced SGS method because one weight captures semantic closeness and the other adjusts for word importance, which avoids a situation where one overpowers the other. By using silhouette score as the optimization criterion, the method quantitatively identifies the weight configuration that produces the most cohesive and well-separated clusters. The clustering, in which each text (e.g., topic) is represented by the vector of its corresponding SGS-based similarity scores with the other texts, is carried out using K-means, and the optimal number of clusters is determined using the elbow method. The process simultaneously optimizes for both the weight parameters and the number of clusters. As a result, this grid search systematically explores the parameter space to identify optimal weights configurations for each of the six datasets used for HAMLET's evaluation. It is worth noting that for all six dataset cases, $weight_{wmd}$ = 0.900 and $weight_{idf}$ = 0.099.

## 4.4. Hybrid Graph Network Construction

At this stage, the initial pool of topics is generated using GPT-4o, the embedding method (i.e., SBERT-BERT Hybridization) is developed, and the similarity approach (i.e., SGS method) is established. These three components enable the construction of a hybrid graph network that integrates documents or texts (i.e., comments or news), topics, and words as different node types in a heterogeneous graph structure. The approach is particularly notable for capturing semantic relationships across different textual types. Figure 8 shows the 3D version of the constructed graph for the case of English dataset 1.

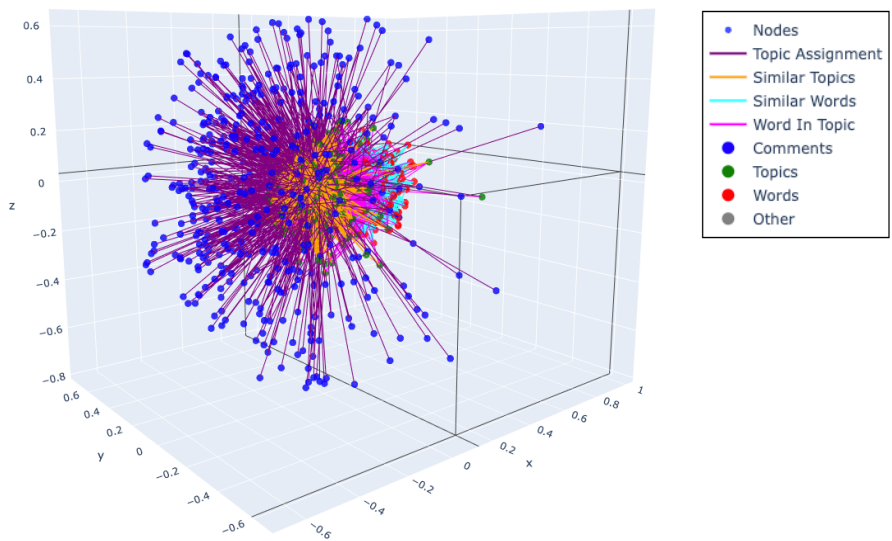

**Figure 8.** Hybrid Graph Network: Case of English Dataset 1

This hybrid graph network is created with three distinct node types:

- **Document nodes:** representing individual text documents, either comments or news
- **Topic nodes:** representing the topics generated by the LLM
- **Word nodes:** representing unique words that form the different topics

Additionally, this graph network is characterized by a rich edge structure that captures four types and similarities:

- **Topic assignment edges:** connecting topics to comments they are assigned to
- **Similar topics edges:** connecting topics that have SGS-based similarity above a threshold
- **Similar words edges:** connecting words that have SGS-based similarity above a threshold
- **Word in topic edges:** connecting words to topics they are associated with

It is worth noting that SGS was employed for both topics and words similarities computations.

For the development of HAMLET for topic modeling, leveraging a hybrid graph network is particularly valuable because it preserves the hierarchical relationship between documents, topics, and words, leverages embeddings to establish meaningful similarity relationships, and more importantly, enables both structural and semantic analysis of the text corpus.

Table 4 summarizes the details of the hybrid graph network for each of the datasets. The thresholds selected reflect the 90th percentile for each case. This choice is justified as an optimal balance that retains only the most meaningful semantic connections while significantly reducing computational complexity. This adaptive threshold is preferable to fixing it to a value that varies between 0.7 and 0.9, as was experimented with, because it ensures consistent graph sparsity across different datasets by preserving statistically significant relationships among topics and words, respectively. These thresholds allow for maintaining an appropriate edge density that balances information retention and computational feasibility, especially as the number of edges changes significantly as the thresholds vary from 0.7 to 0.8 and 0.9.

As outlined below, the French datasets consistently show higher similarity thresholds for both topics and words than English ones, which suggests potentially more semantic coherence in the French corpora. Moreover, French datasets contain significantly more nodes and edges than English datasets.

Table 3. Graph Network Details by Dataset

| Graph Network Details | | English Datasets | | | French Datasets | | |
|---|---|---|---|---|---|---|---|
| | | 1 | 2 | 3 | 1 | 2 | 3 |
| Topics Similarity Threshold | | 0.876 | 0.888 | 0.839 | 0.928 | 0.932 | 0.933 |
| Words Similarity Threshold | | 0.873 | 0.834 | 0.769 | 0.883 | 0.882 | 0.884 |
| Number of Nodes | | 1776 | 691 | 184 | 3159 | 3115 | 3349 |
| Nodes | Comments | 419 | 127 | 23 | 1000 | 999 | 1000 |
| | Topics | 639 | 257 | 59 | 1133 | 1124 | 1233 |
| | Words | 718 | 307 | 102 | 1026 | 992 | 1116 |
| Number of Edges | | 48964 | 9127 | 943 | 122196 | 117998 | 144319 |
| Edges | Topic Assignment | 829 | 306 | 72 | 2599 | 2773 | 2941 |
| | Similar Topics | 20417 | 3303 | 175 | 64185 | 63169 | 76015 |
| | Word in Topic | 1941 | 805 | 175 | 2778 | 2852 | 3090 |
| | Similar Words | 25777 | 4713 | 521 | 52634 | 49204 | 62273 |

### 4.5. EdgeRefineGNN

Once the hybrid graph network is constructed, an edge-conditioned GNN, designed to process heterogeneous semantic networks that contain documents, topics, and words as different node types, is implemented. This neural network, EdgeRefineGNN, leverages convolution operations in a multi-layered manner to refine topic embeddings by aggregating information from connected nodes while considering edge types.

### 4.5.1. Architecture Overview

The architecture of EdgeRefineGNN is characterized first by node processing, where independent dedicated embedding layers process each node type to transform the initial dimension of the feature vectors into a common hidden dimension. Subsequently, three layers of edge-conditioned convolution that allows message passing between nodes while considering edge features (i.e., edge type and similarity scores) are used. Each convolution layer is followed by a normalization layer to stabilize training, a Rectified Linear Unit (ReLU) for computationally efficient complex pattern learning, and a dropout layer (i.e., rate = 0.2) to prevent overfitting. In addition to the normalization and regularization, the architecture incorporates residual connections that start from the second layer to enable better gradient flow and information preservation. Lastly, the topic refinement component is reached, in which the focus is directed to the topic nodes' new embeddings that are transformed back to match the dimensionality of the original topic embeddings, which allows for a direct comparison between the refined and original embeddings.

### 4.5.2. Message Passing Mechanism

The core of EdgeRefineGNN relies on the message passing mechanism:

- **Edge-conditioned Weight Generation:** The model uses a Multilayer Perceptron (MLP) $f_\theta$ to transform the edge features $e_{ij}$ (i.e., type and similarity) into edge-specific weight matrix $W_{ij}$:

$$W_{ij} = f_\theta(e_{ij})$$

- **Message Function:** The messages between nodes are computed as $m_{ij}$ where the node j's embeddings, $h_j$, is transformed by the edge-specific weight matrix before being passed to node i:

$$m_{ij} = W_{ij} h_j$$

- **Message Aggregation:** The model resorts to mean pooling to aggregate messages from all neighboring nodes at layer l before feeding it to layer l+1:

$$h_i^{(l+1)} = \frac{1}{|\mathcal{N}(i)|} \sum_{j \in \mathcal{N}(i)} m_{ij}, \text{ for all neighboring } j \in \mathcal{N}(i)$$

- **Complete Layer Update:** The final node update also incorporates layer normalization and a nonlinear ReLU activation function σ, a dropout layer, in addition to a residual connection from the previous layer:

$$h_i^{(l+1)} = \text{Dropout}\left(\sigma\left(\text{LayerNorm}\left(\frac{1}{|\mathcal{N}(i)|} \sum_{j \in \mathcal{N}(i)} f_\theta(e_{ij}) h_j + h_i^{(l)}\right)\right)\right)$$

### 4.5.3. Loss Function

The model is trained using Mean Squared Error (MSE) loss:

$$\text{Loss} = \frac{1}{N} \sum_{i=1}^{N} (y_i - \hat{y}_i)^2$$

where
- $y_i$ is the original topic embeddings from the input graph
- $\hat{y}_i$ is the refined topic embeddings output by the EdgeRefineGNN
- N is the number of topic nodes

This loss function ensures that refined topic embeddings maintain similarity to their original representations while being enhanced by the graph's structural information.

The training process focuses on refining topic representations in which the MSE loss measures the quality of topic embeddings after message passing. An Adam optimizer is employed with a learning rate of 0.001, and EdgeRefineGNN is trained for 100 epochs.

EdgeRefineGNN effectively captures the complex interdependencies between documents, topics, and words, improving the downstream topic modeling. This is achieved by maintaining edge type-specific processing while enabling information exchange across heterogeneous node types, where edge features are leveraged to condition the message passing based on relationship types and similarities. The goal is to refine the topic embeddings while preserving original semantic content.

## 4.6. Top k Topics Extraction

After refining the topic embeddings using EdgeRefineGNN, clustering is carried out using the K-means method. The number of clusters reflects the number k of top topics targeted for extraction. The value of k depends on the user; however, for the datasets used in this research, it is set to 50, except for English Dataset 3 where it was chosen to be 25 due to its smaller size compared to the others. When implementing K-means, 10 different initializations were performed to find optimal clustering. The optimal clustering is determined using the inertia (i.e., within-cluster sum-of-squares), in which lower values indicate better results.

$$\text{inertia} = \sum \min(||\text{emb}_t - \mu||^2)$$

where

- $\text{emb}_t$ is the embedding of the topic t
- $\mu$ is the centroid of the cluster to which the topic t belongs

The topics are organized into clusters with the minimum cluster size set to 1 to ensure meaningful groupings. The clustering approach serves as the foundation for the representative topic selection process:

- **Coherence-based Extraction:** the coherence-based approach evaluates how well each topic's embedding aligns with its associated words, focusing on semantic quality:

$$\text{Coherence}_t = \text{avg}(\text{cosine\_sim}(\text{emb}_t, \text{emb}_{\text{neighboring\_word}}))$$

- **Centroid-based Extraction:** the centroid-based approach focuses on topic identification that are closest to the geometric center of their cluster in the embedding space:

$$\text{Centroid}_t = \text{cosine\_sim}(\text{emb}_t, \text{centroid})$$

- **Connectivity-based Extraction:** the connectivity-based approach examines the structural connections between topics and words in the graph:

$$\text{Connectivity}_t = |\text{word\_connections}| \times \text{avg}(\text{word\_similarities})$$

Given that HAMLET is an unsupervised approach, it would be possible to employ all three extraction methods and select the one that yields the highest performance score before deducing the top k topics.

One of the main advantages that HAMLET brings to the table is the format of the extracted final topics. Unlike existing topic models where each topic is represented by a bag-of-words and human interpretation is required to deduce and name the topics, HAMLET's topics are short phrases that can directly be used as labels for healthcare text classification applications (i.e. Appendix B.2).

## 4.7. Performance Metrics

To assess the performance of the different topic models (i.e., HAMLET and existing methods), various metrics are leveraged. For traditional models (i.e., Latent Dirichlet Allocation (LDA), Latent Semantic Analysis (LSA), Non-negative Matrix Factorization (NMF), and BERTopic), four conventional metrics were used. They include $C_v$ coherence that aligns with human judgment by measuring semantic similarity among top words in a topic, $C_{npmi}$ coherence that focuses on word co-occurrence patterns using normalized pointwise mutual information, Topic Diversity that quantifies how unique topics are across a model and Mean Pairwise Jaccard Similarity that measures topic overlap by averaging the ratio of shared words between topic pairs. However, because the topics generated by existing topic models are represented by bag-of-words, while those that result from HAMLET are short phrases, it would not be relevant to use the aforementioned conventional metrics to evaluate the proposed approach. For this purpose, this research introduces a new metric, the Composite Score, designed to evaluate HAMLET by leveraging five existing topic modeling metrics.

### 4.7.1. Topic Diversity
Topic Diversity measures how varied the topics are by calculating the ratio of unique words across all topics to the total word count:

$$\text{Topic Diversity} = \frac{|U|}{|W|}$$

where
- U is the set of unique words across all topics
- W is the total count of words across all topics

Higher values indicate greater diversity among topics. This metric is important because it quantifies how distinct topics are. If the topic diversity is high, the model has identified a wide range of unique topics rather than repeating similar terms across multiple topics.

### 4.7.2. Mean Pairwise Jaccard Similarity
Mean Pairwise Jaccard Similarity measures the overlap between topics in terms of shared words:

$$\text{Mean Pairwise Jaccard Similarity} = \frac{1}{\binom{N}{2}} \sum_{i=1}^{N-1} \sum_{j=i+1}^{N} \frac{|T_i \cap T_j|}{|T_i \cup T_j|}$$

where
- $T_i$ and $T_j$ are the sets of words from two different topics
- N is the total number of topics

Lower values are the target in this case, because they indicate less overlap between topics. This metric reflects topic redundancy. When the Mean Pairwise Jaccard Similarity is high, multiple topics may capture the same underlying concepts, which should be prevented.

### 4.7.3. Topic Coherence
Topic Coherence measures how semantically coherent the topics are. For the case of HAMLET's evaluation, $\text{Score}_{\text{coherence}_i}$ is the average of the similarities between topic i and its neighboring words:

$$\text{Coherence} = \frac{1}{N} \sum_{i=1}^{N} \text{Score}_{\text{coherence}_i}$$

Higher values indicate more semantically coherent topics. This metric evaluates whether the words represent a topic make sense together through cosine similarity. Unlike the two previous metrics that focus on the distinction between topics, coherence focuses on the quality of each topic to determine whether the words within a topic truly belong together semantically.

### 4.7.4. Silhouette Score
Conventionally, the Silhouette Score (Rousseeuw, 1987) measures the quality of clustering by evaluating the separation between topic clusters:

$$\text{Silhouette Score} = \frac{1}{N} \sum_{i=1}^{N} \frac{b_i - a_i}{\max(a_i, b_i)}$$

where
- $a_i$ is the average distance from i to all other points in its cluster
- $b_i$ is the average distance from i to points in the nearest different cluster

The Silhouette Score here evaluates the topic model quality by measuring how distinct and well-separated topic embeddings are in a semantic space. When applied to vectorized topics, a higher score indicates that each topic represents a unique concept with clear boundaries between different topics. Therefore, it helps assess whether HAMLET has successfully identified diverse, non-redundant topics that effectively capture different aspects of the corpus.

### 4.7.5. Davies-Bouldin Score

Conventionally, the Davies-Bouldin Score (Davies and Bouldin, 1979) measures cluster compactness and separation:

$$\text{Davies-Bouldin Score} = \frac{1}{N} \sum_{i=1}^{N} \max_{j \neq i} \frac{\sigma_i + \sigma_j}{d_{ij}}$$

where
- $\sigma_i$ is the average intra-cluster distance for cluster i
- $d_{ij}$ is the distance between centroids of clusters $i$ and $j$

In this case, the Davies-Bouldin Score measures the average similarity between each topic embedding and its most similar neighbor, relative to the distance between them. A lower score indicates greater separation between individual topic embeddings, which means the topics represent more distinct concepts with minimal semantic overlap. This also helps evaluate whether the model has identified a diverse set of topics that cover different aspects of the corpus without redundancy.

### 4.7.6. Composite Score

The Composite Score combines all the metrics above to provide a comprehensive evaluation:

$$\text{Composite Score} = 0.4 \times \text{Topic Diversity} + 0.15 \times \text{Jaccard Similarity} + 0.2 \times \text{Coherence} + 0.2 \times \text{Silhouette Score} + 0.05 \times \text{Davies-Bouldin Score}$$

All the metrics are normalized to a 0-1 scale, in which higher values are better. The Composite Score provides a balanced assessment of HAMLET's topics' quality by weighting different aspects of performance. The weights reflect the relative importance of each metric as well as its stability based on the sensitivity analysis conducted (see Figures 9 to 14). Topic Diversity was given the highest weight (w = 0.4), which highlights the importance of having distinct topics in a good topic model. This highest weight is also justified by its exceptional stability across all datasets and consistent performance under perturbations. Jaccard Similarity's weight (w = 0.15) is limited due to high volatility, but it is kept non-negligible to maintain its valuable contribution. Coherence and Silhouette's weights (w = 0.2) are balanced, which reflect their moderate sensitivity and reliable behavior. Coherence, despite its more pronounced sensitivity, remains a key metric. Davies-Bouldin's weight (w = 0.05) is low due to its inconsistent behavior across datasets and for not being a primary evaluation metric.

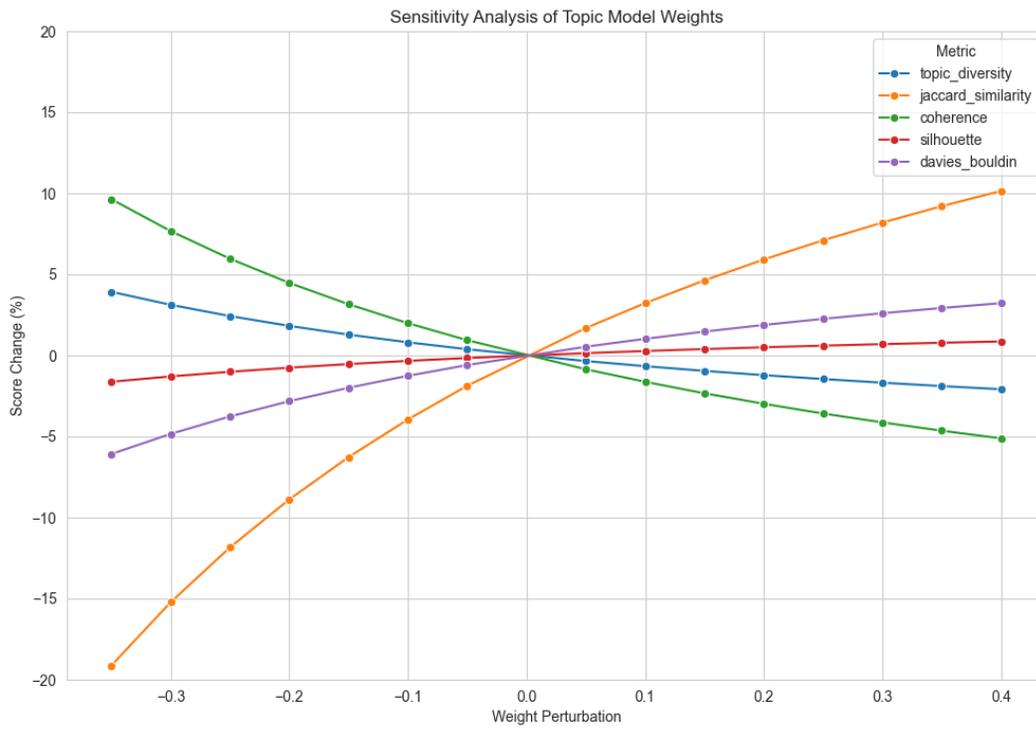

**Figure 9.** Composite Score Sensitivity Analysis: Case of English Dataset 1

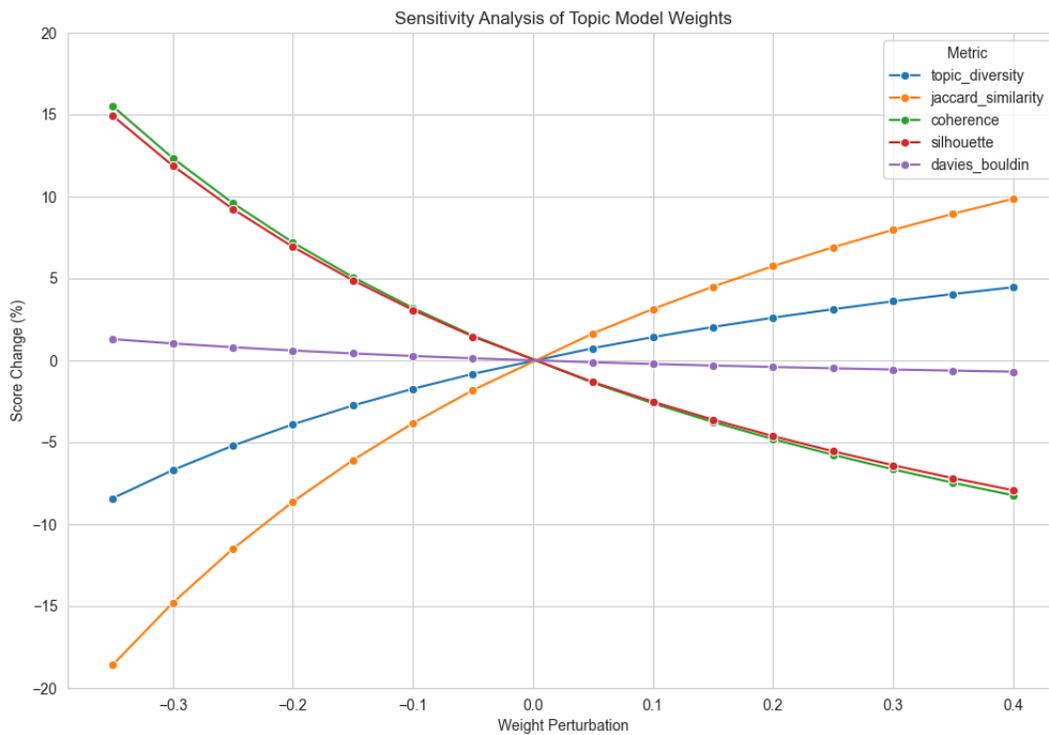

**Figure 10.** Composite Score Sensitivity Analysis: Case of English Dataset 2

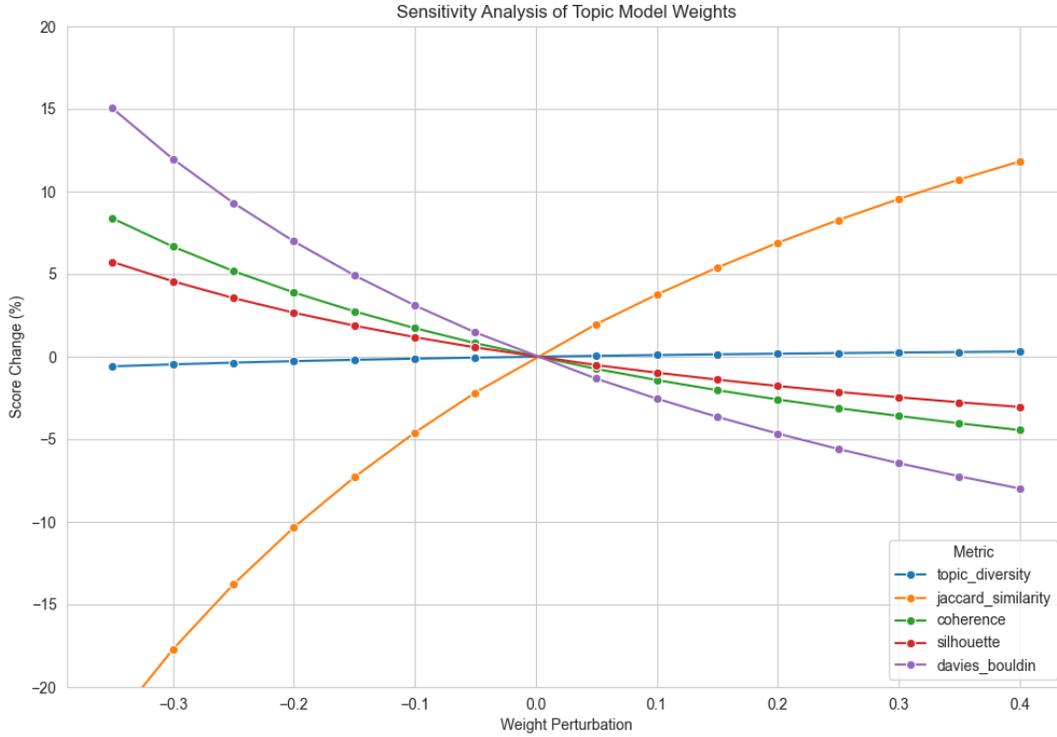

**Figure 11.** Composite Score Sensitivity Analysis: Case of English Dataset 3

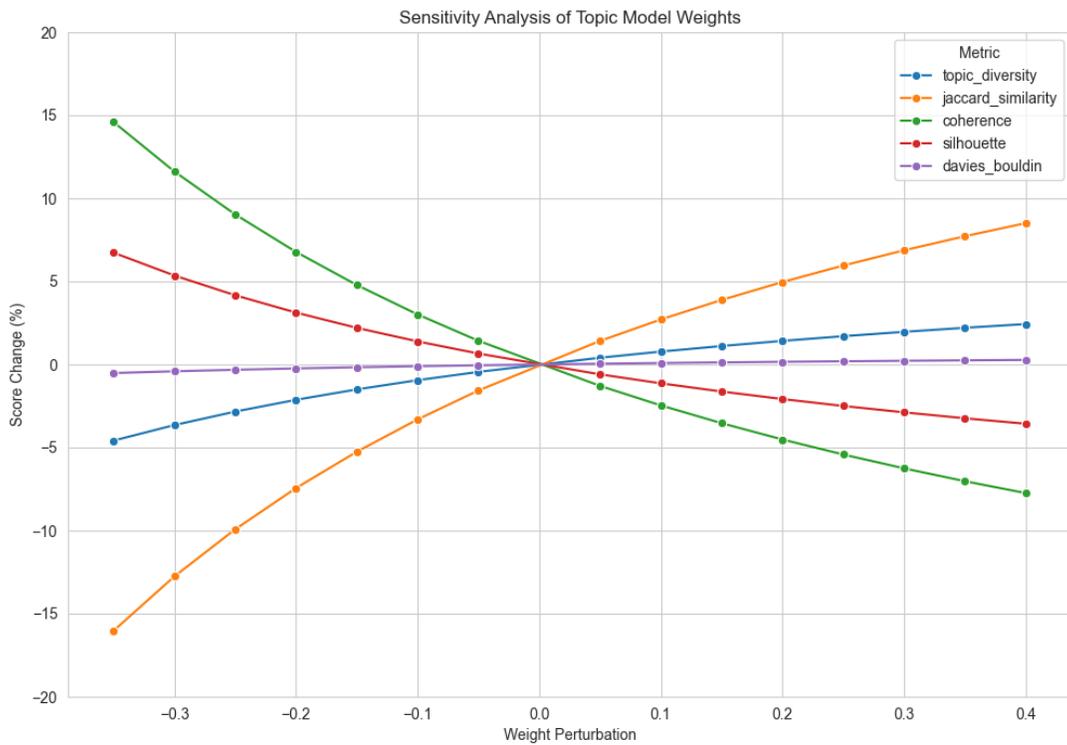

**Figure 12.** Composite Score Sensitivity Analysis: Case of French Dataset 1

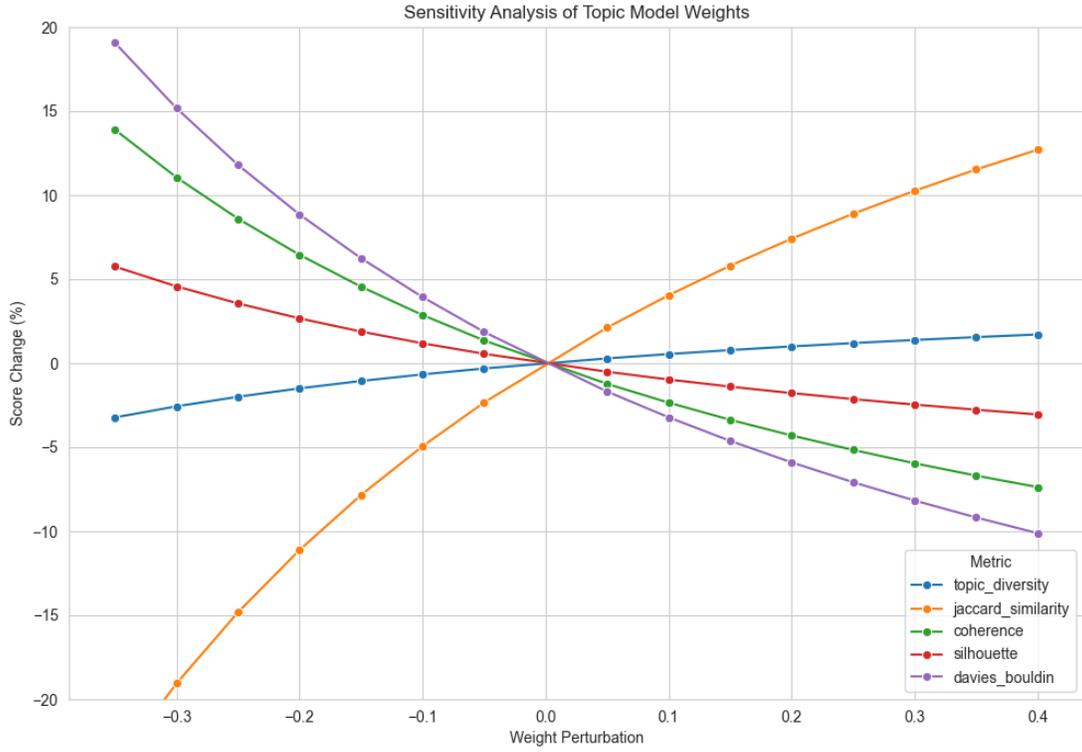

**Figure 13.** Composite Score Sensitivity Analysis: Case of French Dataset 2

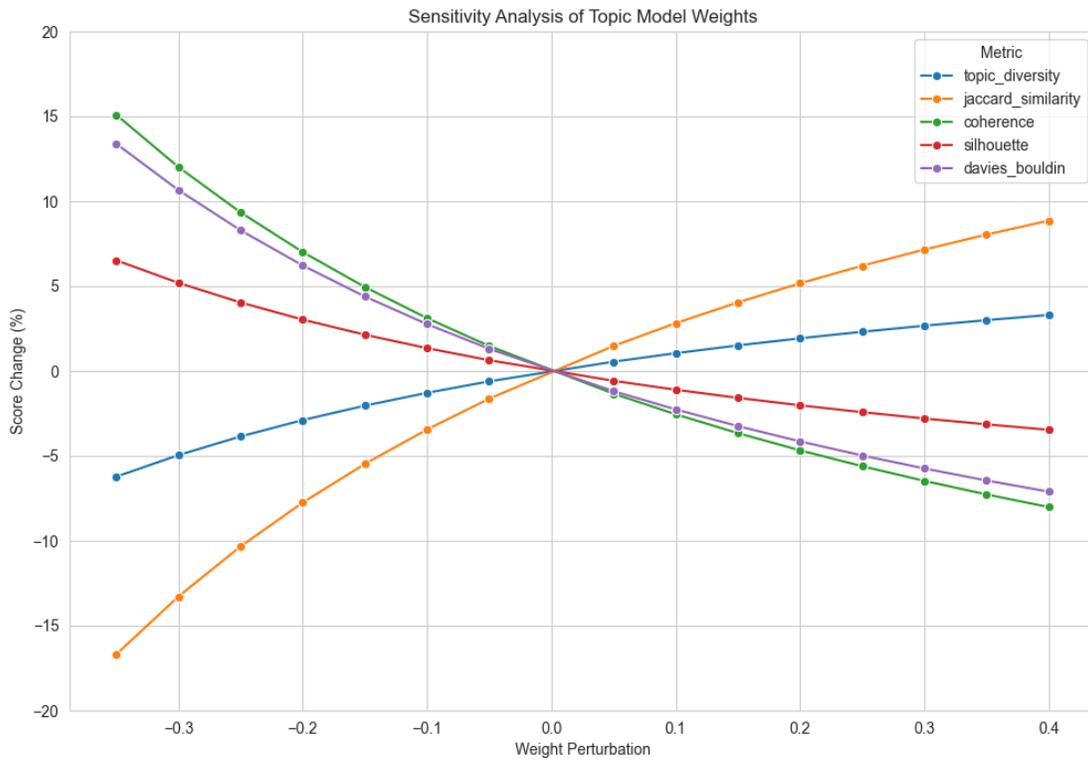

**Figure 14.** Composite Score Sensitivity Analysis: Case of French Dataset 3

# 5. Results and Discussion

In this section, HAMLET's performance is discussed for the case of three English datasets and three French datasets. Existing topic models are first used and evaluated. Afterward, HAMLET is assessed through the proposed Composite Score. Because the conventional methods and the proposed approach operate differently, a direct comparison is irrelevant. However, HAMLET already has an edge given the format of the output topics that enables an automated end-to-end topic modeling, which does not require human assistance and intervention for interpretation, as is the case of existing topic models. Furthermore, a statistical validation is carried out to ensure HAMLET's repeatability. Additionally, an ablation study is conducted to showcase the importance of the EdgeRefineGNN component by comparing the performance when using the refined embeddings to the original embeddings.

## 5.1. Topic Modeling Existing Methods

The existing topic models experimented with include LDA, LSA, NMF, and BERTopic. The comparisons conducted highlight the performance differences across languages as well as the dataset's average document lengths.

### 5.1.1. English Datasets

Tables 5-7 reveal interesting patterns about how different topic models perform across the three English datasets. In terms of $C_v$, LDA shows consistent improvement in coherence (0.600 → 0.642 → 0.699) as document length increases, which suggests that more contextual information is beneficial, while LSA, NMF, and BERTopic show modest improvements in $C_v$ across datasets compared to LDA. Regarding $C_{npmi}$, LDA shows a decline (0.021 → -0.233 → -0.255), which indicates that despite improved $C_v$, as the dataset's documents get longer, the topics become less meaningful. However, NMF shows an improvement (-0.263 → -0.090 → -0.042) as document length increases, which means that it handles longer texts better than shorter ones from this metric's perspective. LSA does not show any consistency as the average number of tokens of the dataset changes. BERTopic maintains relatively stable values. Concerning Topic Diversity, NMF consistently maintains the highest topic diversity across all datasets (0.662, 0.672, 0.656), which indicates that it generates the most distinct topics. LDA and BERTopic show inconsistent topic diversity across datasets while LSA demonstrates improvements as the document length increases (0.198 → 0.264 → 0.318). When it comes to Mean Pairwise Jaccard Similarity, LDA shows inconsistent yet high similarity, which suggests potential topic redundancy despite its high $C_v$ scores. LSA and NMF maintain consistently low similarities, which aligns with the high diversity for the case of NMF. BERTopic reveals an increasing similarity score as the documents become lengthier.

**Table 4.** Existing Topic Models Results – English Dataset 1

| English Dataset 1 | $C_v$ | $C_{npmi}$ | Topic Diversity | Mean Pairwise Jaccard Similarity |
|---|---|---|---|---|
| LDA | 0.600 | 0.021 | 0.116 | 0.786 |
| LSA | 0.291 | -0.161 | 0.198 | 0.076 |
| NMF | 0.401 | -0.263 | 0.662 | 0.011 |
| BERTopic | 0.354 | -0.044 | 0.263 | 0.159 |

**Table 5.** Existing Topic Models Results – English Dataset 2

| English Dataset 2 | $C_v$ | $C_{npmi}$ | Topic Diversity | Mean Pairwise Jaccard Similarity |
|---|---|---|---|---|
| LDA | 0.642 | -0.233 | 0.330 | 0.427 |
| LSA | 0.328 | -0.058 | 0.264 | 0.055 |
| NMF | 0.422 | -0.090 | 0.672 | 0.014 |
| BERTopic | 0.367 | -0.008 | 0.147 | 0.325 |

**Table 6.** Existing Topic Models Results – English Dataset 3

| English Dataset 3 | $C_v$ | $C_{npmi}$ | Topic Diversity | Mean Pairwise Jaccard Similarity |
|---|---|---|---|---|
| **LDA** | 0.699 | -0.255 | 0.246 | 0.538 |
| **LSA** | 0.341 | -0.082 | 0.318 | 0.052 |
| **NMF** | 0.442 | -0.042 | 0.656 | 0.019 |
| **BERTopic** | 0.367 | -0.008 | 0.184 | 0.365 |

LDA shows the highest $C_v$ but increasingly deteriorating $C_{npmi}$ and high similarity between topics, which suggests that for longer documents, it creates coherent but potentially redundant and less interpretable topics. LSA demonstrates the weakest coherence but maintains modest diversity and low similarity. NMF reveals the best balance across metrics, in which moderate $C_v$, improves $C_{npmi}$, high diversity, and low similarity. Additionally, it appears to handle increasing document length effectively. BERTopic indicates moderate performance across metrics, in which it stabilizes $C_{npmi}$ but decreases diversity as document length increases. The results suggest that for longer documents, NMF offers the most balanced performance when it considers all metrics together.

### 5.1.2. French Datasets

Tables 8-10 summarize the results across the three French datasets with increasing document length for each of the four topic models. In terms of $C_v$, LDA shows an inconsistent pattern (i.e., 0.438 → 0.713 → 0.565), with a significant improvement in Dataset 2 followed by a decline in Dataset 3. LSA demonstrates a declining score as document length increases (i.e., 0.441 → 0.372 → 0.340). NMF maintains consistently high coherence across all datasets (i.e., 0.660, 0.642, 0.649), which outperforms other models in Dataset 1 and Dataset 3. BERTopic reveals a slight decline (i.e., 0.405 → 0.398 → 0.373) as document length increases. Regarding $C_{npmi}$, NMF is the only topic model with positive values across all datasets (i.e., 0.024 → 0.037 → 0.040), which suggests that it generates the most interpretable topics for French healthcare textual data. LDA and LSA show poor and declining $C_{npmi}$. BERTopic shows very modest improvement (i.e., -0.033 → -0.015 → -0.017) as document length increases. Concerning the Topic Diversity, NMF consistently maintains the highest topic diversity across all datasets (i.e., 0.836, 0.828, 0.828), which significantly outperforms other topic models. LSA shows relatively stable diversity (i.e., 0.342 → 0.326 → 0.362). LDA demonstrates increasing improvement (i.e., 0.202 → 0.240 → 0.314), while BERTopic shows a significant decline in diversity (i.e., 0.317 → 0.248 → 0.121) as documents lengthen. When it comes to Mean Pairwise Jaccard Similarity, NMF maintains extremely low similarity across all datasets (i.e., 0.006, 0.008, 0.008), which indicates highly distinct topics. LSA also maintains low similarity (i.e., 0.043, 0.049, 0.045). However, LDA shows consistently high similarity (i.e., 0.596, 0.591, 0.441), in which some improvement in Dataset 3 is shown. Also, BERTopic reveals an increase in similarity (i.e., 0.213 → 0.285 → 0.502), which suggests increasingly redundant topics with longer documents.

**Table 7.** Existing Topic Models Results – French Dataset 1

| French Dataset 1 | $C_v$ | $C_{npmi}$ | Topic Diversity | Mean Pairwise Jaccard Similarity |
|---|---|---|---|---|
| **LDA** | 0.438 | -0.173 | 0.202 | 0.596 |
| **LSA** | 0.441 | -0.157 | 0.342 | 0.043 |
| **NMF** | 0.660 | 0.024 | 0.836 | 0.006 |
| **BERTopic** | 0.405 | -0.033 | 0.317 | 0.213 |

**Table 8.** Existing Topic Models Results – French Dataset 2

| French Dataset 2 | $C_v$ | $C_{npmi}$ | Topic Diversity | Mean Pairwise Jaccard Similarity |
|---|---|---|---|---|
| **LDA** | 0.713 | -0.223 | 0.240 | 0.591 |
| **LSA** | 0.372 | -0.178 | 0.326 | 0.049 |
| **NMF** | 0.642 | 0.037 | 0.828 | 0.008 |
| **BERTopic** | 0.398 | -0.015 | 0.248 | 0.285 |

**Table 9.** Existing Topic Models Results – French Dataset 3

| French Dataset 3 | $C_v$ | $C_{npmi}$ | Topic Diversity | Mean Pairwise Jaccard Similarity |
|---|---|---|---|---|
| **LDA** | 0.565 | -0.224 | 0.314 | 0.441 |
| **LSA** | 0.340 | -0.203 | 0.362 | 0.045 |
| **NMF** | 0.649 | 0.040 | 0.828 | 0.008 |
| **BERTopic** | 0.373 | -0.017 | 0.121 | 0.502 |

NMF outperforms the other topic models across all metrics for French datasets, with high $C_v$, positive $C_{npmi}$, exceptionally high diversity, and minimal topic similarity. It handles French language characteristics most effectively. However, LDA shows inconsistent performance with high similarity between topics and negative $C_{npmi}$ values, which suggests that it does not handle French documents correctly. Additionally, LSA demonstrates declining coherences (i.e., $C_v$ and $C_{npmi}$) with modest diversity and low similarity. Moreover, BERTopic reveals a sharp decline in diversity and a sharp increase in similarity for the case of Dataset 3, which shows that it may not scale well with longer French documents. It can be concluded that for French textual data across varying document lengths, NMF provides the most balanced and consistently strong performance across all four metrics.

The comparison of the English and French datasets results reveals interesting language-specific performance differences among topic models. NMF demonstrates superior performance on French datasets while showing moderate but improving performance on English datasets as the documents become longer. LDA performs better on English datasets in terms of $C_v$ but struggles with topic similarity and declining $C_{npmi}$ values as document length increases for both languages. LSA maintains low topic similarity across both languages but generally shows weak $C_v$. BERTopic shows deteriorating trends with longer documents in both languages, but the degradation is more pronounced in French texts. These differences suggest that language-specific characteristics significantly influence each topic model's performance, in which NMF is particularly well-suited to French texts compared to the other models.

Existing topic modeling approaches (i.e., LDA, NMF, LSA, and BERTopic) encounter several key challenges that limit their application to real-world scenarios. These models often struggle with semantic understanding, which generate topics that are statistically coherent but lack meaningful interpretation, as evidenced by the negative $C_{npmi}$ scores across multiple datasets. Moreover, they frequently generate redundant topics with high similarity scores, particularly LDA, which showed exceptionally high Jaccard Similarity values up to 0.786. Furthermore, performance varies substantially across languages, in which models such as NMF perform well with French documents but more moderately with English, which indicates a strong language dependency and the need for language-agnostic topic models. Additionally, document length significantly impacts model performance, which leads to inconsistent results. Finally, these approaches offer limited interpretability of the resulting topics and, therefore, are not suitable for subsequent use as labels for downstream text classification tasks.

## 5.2. HAMLET

The results of the existing approaches highlight the challenges faced and the need for a more robust, language-agnostic, document-length-insensitive healthcare topic model that does not involve humans in the loop. HAMLET is the developed approach that fills this gap in the literature. Similar to the existing topic model evaluation, HAMLET's assessment contrasts datasets of different languages and document lengths. Additionally, the different previously designed top k topics extraction methods (i.e., coherence-based, centroid-based, connectivity-based) are also compared. The Composite Score introduced is used for all the assessments.

### 5.2.1. English Datasets

Tables 11-13 summarize HAMLET's performance in the case of English datasets. The Topic Diversity increases from Dataset 1 (i.e., 0.691 - 0.750) to Dataset 3 (i.e., 0.772 - 0.807). This higher diversity with longer documents shows that more tokens allow for more distinct topics. The Mean Pairwise Jaccard Similarity remains low across all datasets (i.e., 0.007 - 0.018). However, some overlap increases as documents grow longer, as reflected by the slight increase in Dataset 3. Furthermore, the Coherence decreases as document length increases, which means that while longer documents can cover more topics, the resulting topics become less coherent with the neighboring words. Moreover, both the Silhouette and Davies-Bouldin Scores drop in performance from Dataset 1 to Dataset 2, then the performance

is improved in Dataset 3. Lastly, the Composite Score remains relatively stable across datasets, and the coherence-based extraction method consistently outperforms the other two extraction methods.

Table 10. HAMLET Results – English Dataset 1

| English Dataset 1 | Topic Diversity | Mean Pairwise Jaccard Similarity | Coherence | Silhouette Score | Davies-Bouldin Score | Composite Score |
|---|---|---|---|---|---|---|
| **Refined - Coherence** | 0.691 | 0.015 | 0.680 | 0.516 | 0.383 | **0.752** |
| **Refined - Centroid** | 0.750 | 0.007 | 0.524 | 0.515 | 0.386 | 0.746 |
| **Refined - Connectivity** | 0.723 | 0.009 | 0.518 | 0.516 | 0.378 | 0.733 |

Table 11. HAMLET Results – English Dataset 2

| English Dataset 2 | Topic Diversity | Mean Pairwise Jaccard Similarity | Coherence | Silhouette Score | Davies-Bouldin Score | Composite Score |
|---|---|---|---|---|---|---|
| **Refined - Coherence** | 0.732 | 0.012 | 0.623 | 0.306 | 1.296 | **0.714** |
| **Refined - Centroid** | 0.787 | 0.008 | 0.530 | 0.323 | 1.844 | 0.706 |
| **Refined - Connectivity** | 0.724 | 0.010 | 0.544 | 0.303 | 1.306 | 0.695 |

Table 12. HAMLET Results – English Dataset 3

| English Dataset 3 | Topic Diversity | Mean Pairwise Jaccard Similarity | Coherence | Silhouette Score | Davies-Bouldin Score | Composite Score |
|---|---|---|---|---|---|---|
| **Refined - Coherence** | 0.807 | 0.017 | 0.536 | 0.419 | 0.614 | **0.754** |
| **Refined - Centroid** | 0.772 | 0.018 | 0.485 | 0.440 | 0.608 | 0.732 |
| **Refined - Connectivity** | 0.793 | 0.017 | 0.509 | 0.440 | 0.596 | 0.746 |

These experiments prove that HAMLET performs well across varying document lengths, in which coherence-based extraction is the most effective approach. As the average number of tokens increases, the corresponding documents incorporate more diverse topics but with reduced coherence. Moreover, a nonlinear relationship can be recognized where medium-length documents struggle with appropriate topic modeling compared to very short or longer documents. Also, the Composite Score is characterized by its stability despite the fluctuations of the individual metrics, which means that HAMLET maintains consistent overall performance even when document lengths vary in the case of the English language.

### 5.2.2. French Datasets

Tables 14-16 summarize HAMLET's performance in the case of French datasets. The Topic Diversity is high in Dataset 1 (i.e., 0.711 - 0.835), decreases in Dataset 2 (i.e., 0.710 - 0.744), then increases again in Dataset 3 (i.e., 0.750 - 0.793). Unlike the English datasets, a nonlinear relationship with document length can be remarked. The Mean Pairwise Jaccard Similarity remains consistently low across all datasets (i.e., 0.006 - 0.023). However, a peak can be noticed in Dataset 2 (0.022-0.023), which indicates a higher topic overlap in medium-length documents. The Coherence decreases from Dataset 1 (i.e., 0.480 - 0.515) to Dataset 2 (i.e., 0.413 - 0.432), then increases in Dataset 3 (i.e., 0.469 - 0.541). It is worth noting that Dataset 3 shows the highest coherence scores for the coherence-based approach. The Silhouette Score improves significantly in Dataset 2 (i.e., 0.568 - 0.570) compared to Dataset 1 (0.396-0.402) and Dataset 3 (i.e., 0.400 - 0.403). The Davies-Bouldin Score is best (i.e., lowest) in Dataset 2 (i.e., 0.552 - 0.556) compared to Dataset 1 (i.e., 0.695 - 0.731) and Dataset 3 (i.e., 0.686 - 0.697). Lastly, the Composite Score reaches its highest value in Dataset 1 (i.e., 0.759), decreases in Dataset 2 (0.724), and then slightly recovers in Dataset

3 (i.e., 0.748). Once again, the coherence-based method consistently outperforms other extraction methods across all datasets.

Table 13. HAMLET Results – French Dataset 1

| French Dataset 1 | Topic Diversity | Mean Pairwise Jaccard Similarity | Coherence | Silhouette Score | Davies-Bouldin Score | Composite Score |
|---|---|---|---|---|---|---|
| **Refined - Coherence** | 0.835 | 0.006 | 0.515 | 0.402 | 0.695 | **0.759** |
| **Refined - Centroid** | 0.805 | 0.007 | 0.480 | 0.399 | 0.706 | 0.739 |
| **Refined - Connectivity** | 0.711 | 0.014 | 0.480 | 0.396 | 0.731 | 0.700 |

Table 14. HAMLET Results – French Dataset 2

| French Dataset 2 | Topic Diversity | Mean Pairwise Jaccard Similarity | Coherence | Silhouette Score | Davies-Bouldin Score | Composite Score |
|---|---|---|---|---|---|---|
| **Refined - Coherence** | 0.744 | 0.022 | 0.432 | 0.569 | 0.555 | **0.724** |
| **Refined - Centroid** | 0.716 | 0.022 | 0.413 | 0.568 | 0.556 | 0.708 |
| **Refined - Connectivity** | 0.710 | 0.023 | 0.417 | 0.570 | 0.552 | 0.707 |

Table 15. HAMLET Results – French Dataset 3

| French Dataset 3 | Topic Diversity | Mean Pairwise Jaccard Similarity | Coherence | Silhouette Score | Davies-Bouldin Score | Composite Score |
|---|---|---|---|---|---|---|
| **Refined - Coherence** | 0.793 | 0.007 | 0.541 | 0.403 | 0.691 | **0.748** |
| **Refined - Centroid** | 0.750 | 0.009 | 0.469 | 0.403 | 0.686 | 0.716 |
| **Refined - Connectivity** | 0.776 | 0.007 | 0.478 | 0.400 | 0.697 | 0.728 |

This empirical study demonstrates that HAMLET's performance on French text reveals some language-specific characteristics, in which medium-length documents (i.e., Dataset 2) show superior Silhouette and Davies-Bouldin scores despite lower Coherence. The coherence-based extraction remains the most effective approach across all document lengths, which shows the robustness of this method across languages.

Comparing the HAMLET results between English and French datasets reveals similarities and differences. In both languages, the coherence-based extraction method consistently outperformed other methods across all datasets, which demonstrates its language-agnostic effectiveness. However, when it comes to increasing the average document length, the results differed between languages. English datasets show a linear increase in Topic Diversity with longer documents but decreased Coherence, while French exhibited a nonlinear pattern in several metrics. Interestingly, medium-length French documents (Dataset 2) displayed superior clustering metrics (Silhouette and Davies-Bouldin Scores) despite lower Coherence scores, which contrasts with English Dataset 2, which showed the poorest clustering performance. Despite these differences, the Composite Scores remained relatively stable across both languages, which indicates HAMLET's overall robustness across different language.

## 5.3. Ablation Study

An ablation study is conducted to demonstrate the important role that the edge-conditioned GNN plays in HAMLET's design by comparing the Composite Scores obtained when using the original topics' embeddings instead of the refined ones prior to the top k topics extraction regardless of the extraction method.

### 5.3.1. English Datasets

The performance comparison across the three English datasets with increasing document token lengths, shown in Tables 17-20, reveals that the refined embeddings-based approaches consistently outperform the original embeddings-based ones, in which the coherence-based extraction method achieves the highest overall composite score (i.e., 0.740 ± 0.023). While maintaining comparable Topic Diversity, using the refined embeddings substantially improves the Coherence and significantly improves the clustering quality. The refined embeddings with the coherence-based extraction approach emerges as the best performing method. Additionally, these results demonstrate that the topics' refinement, and therefore the leveraging of the edge-conditioned GNN, is a robust improvement to HAMLET's topic modeling regardless of the specific extraction method employed.

Table 17. Performance Comparison of Approaches – English Dataset 1

| English Dataset 1 | Topic Diversity | Mean Pairwise Jaccard Similarity | Coherence | Silhouette Score | Davies-Bouldin Score | Composite Score |
|---|---|---|---|---|---|---|
| **Original - Coherence** | 0.738 | 0.014 | 0.526 | -0.076 | 1.475 | 0.654 |
| **Refined - Coherence** | 0.691 | 0.015 | 0.680 | 0.516 | 0.383 | **0.752** |
| **Original - Centroid** | 0.789 | 0.007 | 0.397 | -0.052 | 2.168 | 0.635 |
| **Refined - Centroid** | 0.750 | 0.007 | 0.524 | 0.515 | 0.386 | 0.746 |
| **Original - Connectivity** | 0.723 | 0.009 | 0.381 | -0.133 | 1.745 | 0.607 |
| **Refined - Connectivity** | 0.723 | 0.009 | 0.518 | 0.516 | 0.378 | 0.733 |

Table 18. Performance Comparison of Approaches – English Dataset 2

| English Dataset 2 | Topic Diversity | Mean Pairwise Jaccard Similarity | Coherence | Silhouette Score | Davies-Bouldin Score | Composite Score |
|---|---|---|---|---|---|---|
| **Original - Coherence** | 0.731 | 0.012 | 0.505 | 0.057 | 1.196 | 0.668 |
| **Refined - Coherence** | 0.732 | 0.012 | 0.623 | 0.306 | 1.296 | **0.714** |
| **Original - Centroid** | 0.764 | 0.008 | 0.437 | 0.129 | 1.146 | 0.676 |
| **Refined - Centroid** | 0.787 | 0.008 | 0.530 | 0.323 | 1.844 | 0.706 |
| **Original - Connectivity** | 0.724 | 0.010 | 0.432 | 0.115 | 0.976 | 0.661 |
| **Refined - Connectivity** | 0.724 | 0.010 | 0.544 | 0.303 | 1.306 | 0.695 |

Table 19. Performance Comparison of Approaches – English Dataset 3

| English Dataset 3 | Topic Diversity | Mean Pairwise Jaccard Similarity | Coherence | Silhouette Score | Davies-Bouldin Score | Composite Score |
|---|---|---|---|---|---|---|
| **Original - Coherence** | 0.818 | 0.015 | 0.378 | -0.029 | 1.262 | 0.666 |
| **Refined - Coherence** | 0.807 | 0.017 | 0.536 | 0.419 | 0.614 | **0.754** |
| **Original - Centroid** | 0.810 | 0.016 | 0.321 | -0.069 | 1.026 | 0.653 |
| **Refined - Centroid** | 0.772 | 0.018 | 0.485 | 0.440 | 0.608 | 0.732 |
| **Original - Connectivity** | 0.793 | 0.017 | 0.371 | -0.063 | 1.163 | 0.653 |
| **Refined - Connectivity** | 0.793 | 0.017 | 0.509 | 0.440 | 0.596 | 0.746 |

Table 20. Performance Comparison of Approaches – English Overall

| English Dataset (Overall) | Topic Diversity | Mean Pairwise Jaccard Similarity | Coherence | Silhouette Score | Davies-Bouldin Score | Composite Score |
|---|---|---|---|---|---|---|
| **Original - Coherence** | 0.762 | 0.014 | 0.470 | -0.016 | 1.311 | 0.663 ± **0.008** |
| **Refined - Coherence** | 0.744 | 0.014 | 0.613 | 0.414 | 0.764 | **0.740** ± 0.023 |
| **Original - Centroid** | 0.788 | 0.010 | 0.385 | 0.003 | 1.446 | 0.655 ± 0.021 |
| **Refined - Centroid** | 0.770 | 0.011 | 0.513 | 0.426 | 0.946 | 0.728 ± 0.020 |
| **Original - Connectivity** | 0.746 | 0.012 | 0.395 | -0.027 | 1.295 | 0.641 ± 0.030 |
| **Refined - Connectivity** | 0.746 | 0.012 | 0.524 | 0.420 | 0.760 | 0.725 ± 0.027 |

### 5.3.2. French Datasets

The performance comparison across the three French datasets, summarized in Tables 21-24, shows once again that the refined embeddings-based approaches consistently outperform their original embeddings-based counterparts, in which the refined embeddings with coherence-based extraction method achieve the highest overall composite score (i.e.,0.743 ± 0.018). As is the case with the English datasets, topic diversity is maintained at comparable levels, which shows slight improvements in some cases. Here too, the topics' embeddings refinement led to improved Coherences, Silhouette Scores, and Davies-Bouldin Scores. The effectiveness of the EdgeRefineGNN in HAMLET is consistent across all three datasets.

Table 21. Performance Comparison of Approaches – French Dataset 1

| French Dataset 1 | Topic Diversity | Mean Pairwise Jaccard Similarity | Coherence | Silhouette Score | Davies-Bouldin Score | Composite Score |
|---|---|---|---|---|---|---|
| **Original - Coherence** | 0.761 | 0.011 | 0.441 | -0.057 | 1.733 | 0.642 |
| **Refined - Coherence** | 0.835 | 0.006 | 0.515 | 0.402 | 0.695 | **0.759** |
| **Original - Centroid** | 0.730 | 0.013 | 0.400 | -0.035 | 1.509 | 0.629 |
| **Refined - Centroid** | 0.805 | 0.007 | 0.480 | 0.399 | 0.706 | 0.739 |
| **Original - Connectivity** | 0.711 | 0.014 | 0.375 | -0.025 | 1.830 | 0.609 |
| **Refined - Connectivity** | 0.711 | 0.014 | 0.480 | 0.396 | 0.731 | 0.700 |

Table 22. Performance Comparison of Approaches – French Dataset 2

| French Dataset 2 | Topic Diversity | Mean Pairwise Jaccard Similarity | Coherence | Silhouette Score | Davies-Bouldin Score | Composite Score |
|---|---|---|---|---|---|---|
| **Original - Coherence** | 0.706 | 0.024 | 0.333 | -0.100 | 1.366 | 0.601 |
| **Refined - Coherence** | 0.744 | 0.022 | 0.432 | 0.569 | 0.555 | **0.724** |
| **Original - Centroid** | 0.699 | 0.024 | 0.299 | -0.144 | 1.487 | 0.584 |
| **Refined - Centroid** | 0.716 | 0.022 | 0.413 | 0.568 | 0.556 | 0.708 |
| **Original - Connectivity** | 0.710 | 0.023 | 0.306 | -0.121 | 1.284 | 0.597 |
| **Refined - Connectivity** | 0.710 | 0.023 | 0.417 | 0.570 | 0.552 | 0.707 |

Table 23. Performance Comparison of Approaches – French Dataset 3

| French Dataset 3 | Topic Diversity | Mean Pairwise Jaccard Similarity | Coherence | Silhouette Score | Davies-Bouldin Score | Composite Score |
|---|---|---|---|---|---|---|
| **Original - Coherence** | 0.783 | 0.006 | 0.439 | -0.101 | 1.326 | 0.657 |
| **Refined - Coherence** | 0.793 | 0.007 | 0.541 | 0.403 | 0.691 | **0.748** |
| **Original - Centroid** | 0.754 | 0.008 | 0.350 | -0.100 | 1.504 | 0.623 |
| **Refined - Centroid** | 0.750 | 0.009 | 0.469 | 0.403 | 0.686 | 0.716 |
| **Original - Connectivity** | 0.776 | 0.007 | 0.351 | -0.100 | 1.318 | 0.637 |
| **Refined - Connectivity** | 0.776 | 0.007 | 0.478 | 0.400 | 0.697 | 0.728 |

Table 24. Performance Comparison of Approaches – French Overall

| French Dataset (Overall) | Topic Diversity | Mean Pairwise Jaccard Similarity | Coherence | Silhouette Score | Davies-Bouldin Score | Composite Score |
|---|---|---|---|---|---|---|
| **Original - Coherence** | 0.750 | 0.014 | 0.404 | -0.086 | 1.475 | 0.633 ± 0.029 |
| **Refined - Coherence** | 0.791 | 0.011 | 0.496 | 0.458 | 0.647 | **0.743** ± 0.018 |
| **Original - Centroid** | 0.728 | 0.015 | 0.350 | -0.093 | 1.500 | 0.612 ± 0.024 |
| **Refined - Centroid** | 0.757 | 0.013 | 0.454 | 0.457 | 0.649 | 0.721 ± 0.016 |
| **Original - Connectivity** | 0.732 | 0.015 | 0.344 | -0.082 | 1.477 | 0.614 ± 0.020 |
| **Refined - Connectivity** | 0.732 | 0.015 | 0.459 | 0.455 | 0.660 | 0.711 ± **0.014** |

## 5.4. Statistical Validation

To verify how reliable HAMLET method is, topic modeling was run five separate times. The approaches cover the three top k topic extraction methods as well as using the topics' refined embeddings compared to original embeddings.

### 5.4.1. English Datasets

The analysis of Composite Score replications across the three English datasets with increasing document lengths, summarized in Tables 25-27, reveals that refined embeddings-based methods consistently outperform original embeddings-based methods by approximately 10 percentage points, in which the coherence-based method achieve the highest mean scores across all datasets, 0.753, 0.706, and 0.730, respectively. The performance indicates a nonlinear relationship with the average number of tokens, as Dataset 2 shows lower scores and higher variability than Datasets 1 and 3. The original embeddings-based approaches reveal more stability across datasets. In contrast, refined embeddings-based methods show greater sensitivity to document length but provide superior Composite Scores.

Table 25. Composite Score Replications Descriptive Statistics – English Dataset 1

| Approach | Mean | Std Dev | Min | Max | 95% CI Lower | 95% CI Upper |
|---|---|---|---|---|---|---|
| **Original - Centroid** | 0.645 | **0.007** | 0.635 | 0.654 | 0.636 | 0.654 |
| **Original - Coherence** | 0.652 | 0.012 | 0.638 | 0.664 | 0.638 | 0.667 |
| **Original - Connectivity** | 0.639 | 0.028 | 0.607 | 0.670 | 0.603 | 0.674 |
| **Refined - Coherence** | **0.753** | 0.018 | 0.727 | 0.778 | 0.730 | 0.776 |
| **Refined - Centroid** | 0.745 | 0.014 | 0.726 | 0.764 | 0.729 | 0.762 |
| **Refined - Connectivity** | 0.749 | 0.023 | 0.718 | 0.775 | 0.720 | 0.777 |

Table 26. Composite Score Replications Descriptive Statistics – English Dataset 2

| Approach | Mean | Std Dev | Min | Max | 95% CI Lower | 95% CI Upper |
|---|---|---|---|---|---|---|
| **Original - Centroid** | 0.634 | 0.028 | 0.598 | 0.676 | 0.599 | 0.6692 |
| **Original - Coherence** | 0.632 | **0.026** | 0.596 | 0.668 | 0.600 | 0.6650 |
| **Original - Connectivity** | 0.614 | 0.030 | 0.580 | 0.661 | 0.576 | 0.6512 |
| **Refined - Coherence** | **0.706** | 0.040 | 0.664 | 0.744 | 0.657 | 0.7552 |
| **Refined - Centroid** | 0.690 | 0.054 | 0.629 | 0.747 | 0.623 | 0.7565 |
| **Refined - Connectivity** | 0.674 | 0.059 | 0.592 | 0.729 | 0.600 | 0.7471 |

Table 27. Composite Score Replications Descriptive Statistics – English Dataset 3

| Approach | Mean | Std Dev | Min | Max | 95% CI Lower | 95% CI Upper |
|---|---|---|---|---|---|---|
| **Original - Centroid** | 0.649 | **0.014** | 0.632 | 0.667 | 0.632 | 0.6655 |
| **Original - Coherence** | 0.655 | 0.016 | 0.632 | 0.671 | 0.635 | 0.6740 |
| **Original - Connectivity** | 0.642 | 0.019 | 0.610 | 0.661 | 0.618 | 0.6664 |
| **Refined - Coherence** | **0.730** | 0.031 | 0.679 | 0.755 | 0.691 | 0.7684 |
| **Refined - Centroid** | 0.727 | 0.042 | 0.661 | 0.765 | 0.675 | 0.7800 |
| **Refined - Connectivity** | 0.719 | 0.034 | 0.662 | 0.746 | 0.677 | 0.7607 |

The t-test results compare five different approaches against the refined embeddings using the coherence-based extraction approach across three English datasets (see Tables 28-30). These comparisons reveal consistent patterns. For all three datasets, the comparisons between original embeddings-based approaches (i.e., Coherence, Connectivity, and Centroid) and the refined embeddings coherence-based extraction approach show statistically significant differences with p-values lower than 0.05. Moreover, the negative mean differences indicate that the refined embeddings using the coherence-based extraction approach consistently outperforms all original embeddings-based approaches. In contrast, no statistically significant differences are observed across all three datasets when comparing this same approach to the other refined embeddings-based approaches (i.e., Centroid and Connectivity). Additionally, the mean differences are minimal. This means that these refined embeddings-based approaches perform similarly. These patterns are consistent across the three datasets with varying document lengths. Therefore, the refined embeddings with coherence-based extraction method significantly outperforms all the original embeddings-based approaches and performs statistically equivalently to other refined embeddings-based approaches.Thereafter, the selection of the extraction method may vary from one dataset to another, which HAMLET easily allows, given the unsupervised nature of this specific component. In parallel, these results highlight the importance of the EdgeRefineGNN component.

Table 28. t-tests Results – English Dataset 1

| Approach | Mean Difference | t-statistic | p-value | Significance |
|---|---|---|---|---|
| Original - Connectivity | -0.115 | -7.565 | 0.000 | True |
| Original - Centroid | -0.108 | -12.279 | 0.000 | True |
| Original - Coherence | -0.101 | -10.347 | 0.000 | True |
| Refined - Centroid | -0.008 | -0.768 | 0.465 | False |
| Refined - Connectivity | -0.005 | -0.360 | 0.728 | False |

Table 29. t-tests Results – English Dataset 2

| Approach | Mean Difference | t-statistic | p-value | Significance |
|---|---|---|---|---|
| Original - Connectivity | -0.092 | -4.153 | 0.003 | True |
| Original - Centroid | -0.074 | -3.476 | 0.008 | True |
| Original - Coherence | -0.072 | -3.312 | 0.011 | True |
| Refined - Centroid | -0.032 | -1.018 | 0.338 | False |
| Refined - Connectivity | -0.016 | -0.552 | 0.596 | False |

Table 16. t-tests Results – English Dataset 3

| Approach | Mean Difference | t-statistic | p-value | Significance |
|---|---|---|---|---|
| Original - Connectivity | -0.087 | -5.330 | 0.001 | True |
| Original - Centroid | -0.081 | -5.344 | 0.001 | True |
| Original - Coherence | -0.075 | -4.825 | 0.001 | True |
| Refined - Centroid | -0.011 | -0.532 | 0.609 | False |
| Refined - Connectivity | -0.002 | -0.106 | 0.918 | False |

The ANOVA results across the three English datasets reveal statistically significant differences among the six compared approaches (see Table 31). However, the strength of these differences varies between datasets. Dataset 1 shows the strongest effect, with a high F-statistic (i.e., 47.969) and effect size (i.e., $\eta^2 = 0.909$). Thereby, 90.9% of the variance in performance can be attributed to the differences between approaches. Dataset 3 demonstrates a moderate effect, with an F-statistic of 11.432 and an effect size of 0.704, which shows that 70.4% of the variance is explained by the approach differences. Dataset 2 exhibits the weakest effect, with a relatively low F-statistic of 3.910 and an effect size of 0.449. This indicates that only 44.9% of the variance is explained by differences between approaches. These results align with the previously discussed ones. Additionally, they confirm that while significant differences exist among approaches across all datasets, the magnitude of these differences varies with the average number of tokens, in which Dataset 2 shows less significant differences than Datasets 1 and 3.

Table 171. ANOVA Results – English Datasets

| Metric | Dataset 1 | Dataset 2 | Dataset 3 |
|---|---|---|---|
| F-statistic | 47.969 | 3.910 | 11.432 |
| p-value | 0.000 | 0.010 | 0.000 |
| $\eta^2$ | 0.909 | 0.449 | 0.704 |

### 5.4.2. French Datasets

The statistical analysis of French datasets, summarized in Tables 32-34, reveals that across three datasets with increasing document length, the refined embeddings coherence-based extraction approach consistently achieves the highest performance with mean scores of 0.705, 0.708, and 0.759 respectively. Thus, it outperforms all other methods. The original embeddings-based approaches show more stability with lower standard deviations compared to the refined embeddings-based approaches. However, the latter achieve superior performance despite higher variability. These findings demonstrate that refined embeddings-based methods perform increasingly better with longer French textual data.

**Table 32.** Composite Score Replications Descriptive Statistics – French Dataset 1

| Approach | Mean | Std Dev | Min | Max | 95% CI Lower | 95% CI Upper |
|---|---|---|---|---|---|---|
| **Original - Centroid** | 0.618 | 0.015 | 0.600 | 0.636 | 0.600 | 0.636 |
| **Original - Coherence** | 0.635 | 0.014 | 0.610 | 0.645 | 0.617 | 0.652 |
| **Original - Connectivity** | 0.617 | **0.010** | 0.609 | 0.633 | 0.604 | 0.629 |
| **Refined - Coherence** | **0.705** | 0.033 | 0.672 | 0.759 | 0.664 | 0.745 |
| **Refined - Centroid** | 0.696 | 0.027 | 0.670 | 0.739 | 0.663 | 0.729 |
| **Refined - Connectivity** | 0.689 | 0.012 | 0.675 | 0.702 | 0.674 | 0.704 |

**Table 33.** Composite Score Replications Descriptive Statistics – French Dataset 2

| Approach | Mean | Std Dev | Min | Max | 95% CI Lower | 95% CI Upper |
|---|---|---|---|---|---|---|
| **Original - Centroid** | 0.619 | 0.027 | 0.584 | 0.658 | 0.586 | 0.653 |
| **Original - Coherence** | 0.632 | 0.023 | 0.601 | 0.663 | 0.603 | 0.661 |
| **Original - Connectivity** | 0.616 | **0.014** | 0.597 | 0.630 | 0.599 | 0.634 |
| **Refined - Coherence** | **0.708** | 0.036 | 0.650 | 0.747 | 0.663 | 0.752 |
| **Refined - Centroid** | 0.694 | 0.039 | 0.629 | 0.728 | 0.645 | 0.742 |
| **Refined - Connectivity** | 0.619 | 0.027 | 0.584 | 0.658 | 0.586 | 0.653 |

**Table 34.** Composite Score Replications Descriptive Statistics – French Dataset 3

| Approach | Mean | Std Dev | Min | Max | 95% CI Lower | 95% CI Upper |
|---|---|---|---|---|---|---|
| **Original - Centroid** | 0.633 | 0.012 | 0.622 | 0.652 | 0.618 | 0.648 |
| **Original - Coherence** | 0.665 | **0.010** | 0.653 | 0.675 | 0.653 | 0.676 |
| **Original - Connectivity** | 0.644 | 0.014 | 0.634 | 0.667 | 0.627 | 0.661 |
| **Refined - Coherence** | **0.759** | 0.032 | 0.724 | 0.796 | 0.719 | 0.799 |
| **Refined - Centroid** | 0.738 | 0.037 | 0.704 | 0.796 | 0.691 | 0.784 |
| **Refined - Connectivity** | 0.735 | 0.031 | 0.704 | 0.769 | 0.697 | 0.773 |

The t-test analysis that compares the refined embeddings coherence-based extraction against the other five approaches across three French datasets reveals consistent and significant performance differences (see Tables 35-37). For all three datasets, the refined embeddings coherence-based extraction significantly outperforms all original embeddings-based approaches (i.e., Coherence, Connectivity, and Centroid) with p-values below 0.05; this indicates strong statistical significance. The mean differences between the refined embeddings coherence-based extraction method and the original embeddings-based approaches suggest that the proposed approach's advantage increases with document length. However, when compared to other refined embeddings-based approaches (i.e., Centroid and Connectivity), the differences are less and not statistically significant. This indicates that while all refined embeddings-based approaches perform similarly, they collectively significantly improve over original embeddings-based approaches as is the case of English Datasets.

**Table 35.** t-tests Results – French Dataset 1

| Approach | Mean Difference | t-statistic | p-value | Significance |
|---|---|---|---|---|
| **Original - Connectivity** | -0.088 | -5.778 | 0.000 | True |
| **Original - Centroid** | -0.087 | -5.436 | 0.001 | True |
| **Original - Coherence** | -0.070 | -4.424 | 0.002 | True |
| **Refined - Centroid** | -0.016 | -1.026 | 0.335 | False |
| **Refined - Connectivity** | -0.009 | -0.458 | 0.659 | False |

Table 36. t-tests Results – French Dataset 2

| Approach | Mean Difference | t-statistic | p-value | Significance |
|---|---|---|---|---|
| Original - Connectivity | -0.091 | -5.291 | 0.001 | True |
| Original - Centroid | -0.088 | -4.407 | 0.002 | True |
| Original - Coherence | -0.076 | -3.958 | 0.004 | True |
| Refined - Centroid | -0.017 | -0.740 | 0.480 | False |
| Refined - Connectivity | -0.014 | -0.581 | 0.577 | False |

Table 37. t-tests Results – French Dataset 3

| Approach | Mean Difference | t-statistic | p-value | Significance |
|---|---|---|---|---|
| Original - Connectivity | -0.126 | -8.175 | 0.000 | True |
| Original - Centroid | -0.115 | -7.330 | 0.000 | True |
| Original - Coherence | -0.095 | -6.298 | 0.000 | True |
| Refined - Centroid | -0.024 | -1.218 | 0.258 | False |
| Refined - Connectivity | -0.021 | -0.959 | 0.366 | False |

The ANOVA results for the French datasets also demonstrate strong statistical evidence for significant differences among the six approaches across all three datasets, as shown in Table 38. All datasets show highly significant results with p-values of 0.000. Thus, the observed differences between approaches are not due to chance. Dataset 3 shows the highest F-value, which suggests that the differences between approaches become more significant with longer documents. The effect size is large across all datasets. This means that a substantial percentage of the variance is explained by the approach used. Dataset 3 shows the largest effect size (i.e., 0.830), followed by Dataset 1 (i.e., 0.812) and then Dataset 2 (i.e., 0.658).

Table 38. ANOVA Results – French Datasets

| Metric | Dataset 1 | Dataset 2 | Dataset 3 |
|---|---|---|---|
| F-statistic | 20.768 | 9.233 | 23.406 |
| p-value | 0.000 | 0.000 | 0.000 |
| $\eta^2$ | 0.812 | 0.658 | 0.830 |

It is worth noting that the boxplots that correspond to each of the six comparisons can be found in Appendix B.1.

## 6. Conclusion and Future Directions

This research introduces HAMLET, a novel framework that addresses the challenge of healthcare text classification when labeled data is missing. HAMLET combines LLMs for topic generation with GNNs to refine these topics' embeddings that subsequently serve as classification labels. The framework's core innovations include an SBERT-BERT hybridization approach that enhances both text documents and topic embeddings and a Semantic-Geometric Similarity (SGS) method that improves similarity computations among words and topics. Moreover, the GNN architecture leverages edge-specific features to create more coherent and meaningful topics. A key strength of HAMLET is its ability to maintain interpretability while delivering strong performance. Additionally, the final topics generated do not require further human involvement or topic naming, which is particularly valuable in healthcare applications.

Among the future research directions that are worth exploring is the adaptation of HAMLET to other domains beyond healthcare. This expansion will allow for the validation of the proposed approach as well as its refinement for more generalization. Moreover, because HAMLET is a topic model, one important parameter that is selected is k, the number of topics to extract. Therefore, another interesting research line would be to investigate the integration of methods such as the Hierarchical Dirichlet Process (HDP) with HAMLET for an automated selection.

HAMLET demonstrates the potential of combining Generative AI capabilities with structural architectures to address the limitations of existing topic model approaches in label-scarce environments such as healthcare.

## Appendix A: Data Sources

Data used for this research is publicly available and their corresponding resources are as follows:

- Patient Comments: [NHS England, NHS Choices, Hospitals, Patient Comments and Ratings](#), May 2014

- French Covid-19 News: [gustavecortal/fr_covid_news](#), HuggingFace, January 2025

## Appendix B: Supplementary Materials
### B.1. Statistical Validation – Boxplot

#### B.1.1. English Dataset 1

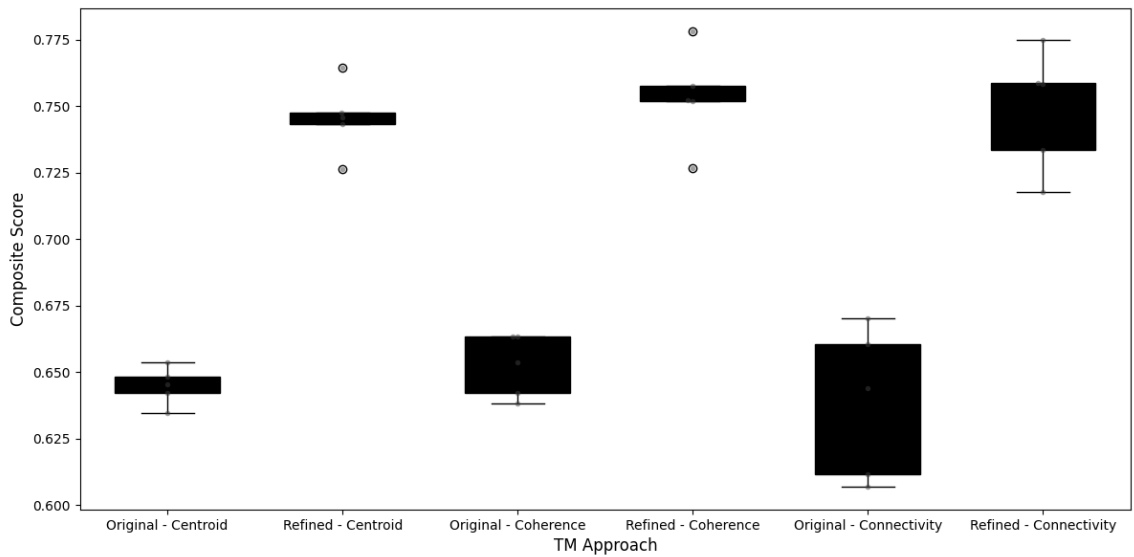

#### B.1.2. English Dataset 2

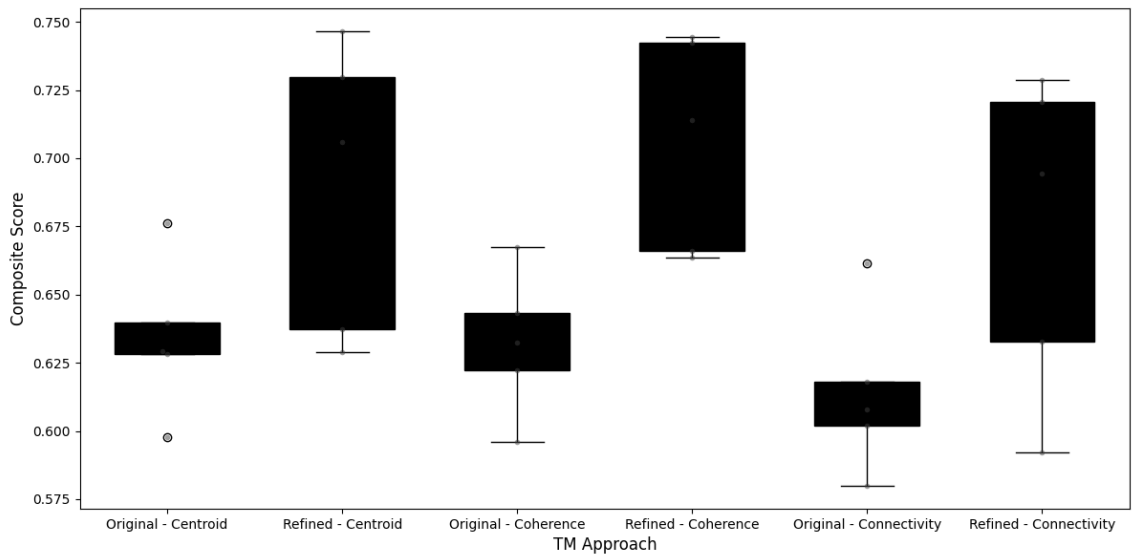

### B.1.3. English Dataset 3

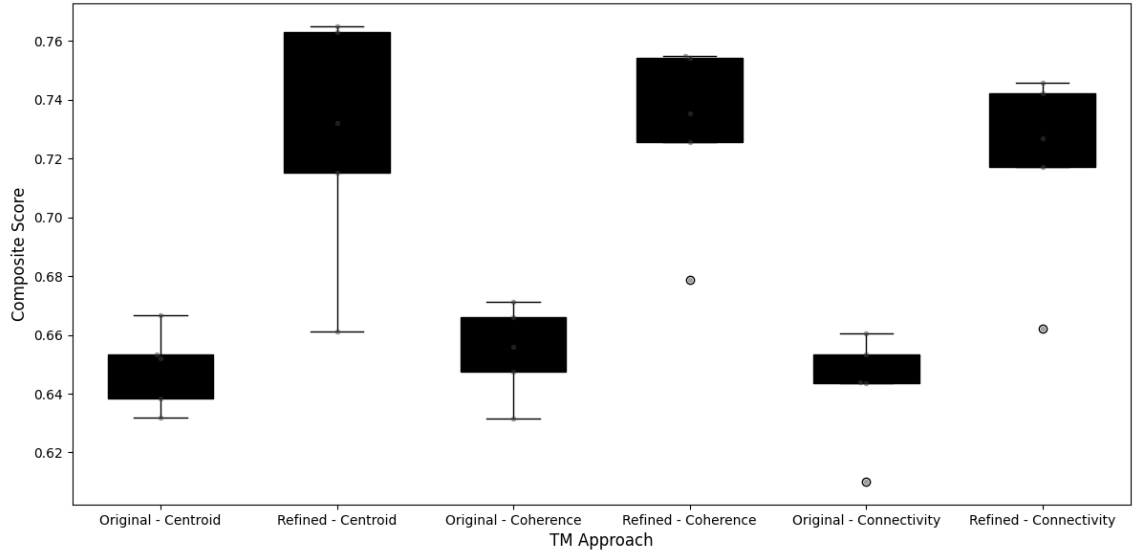

### B.1.4. French Dataset 1

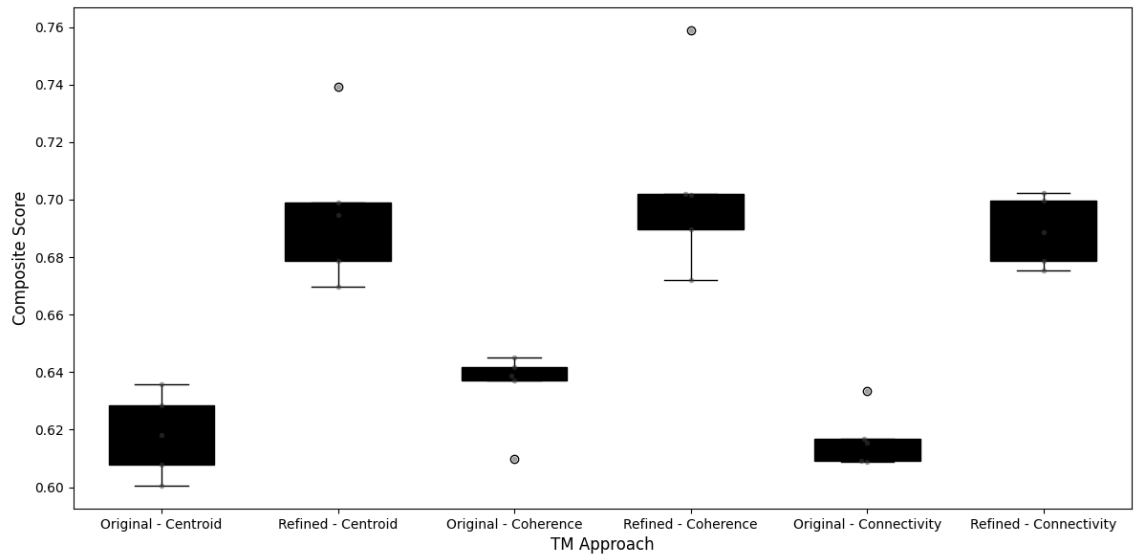

## B.1.5. French Dataset 2

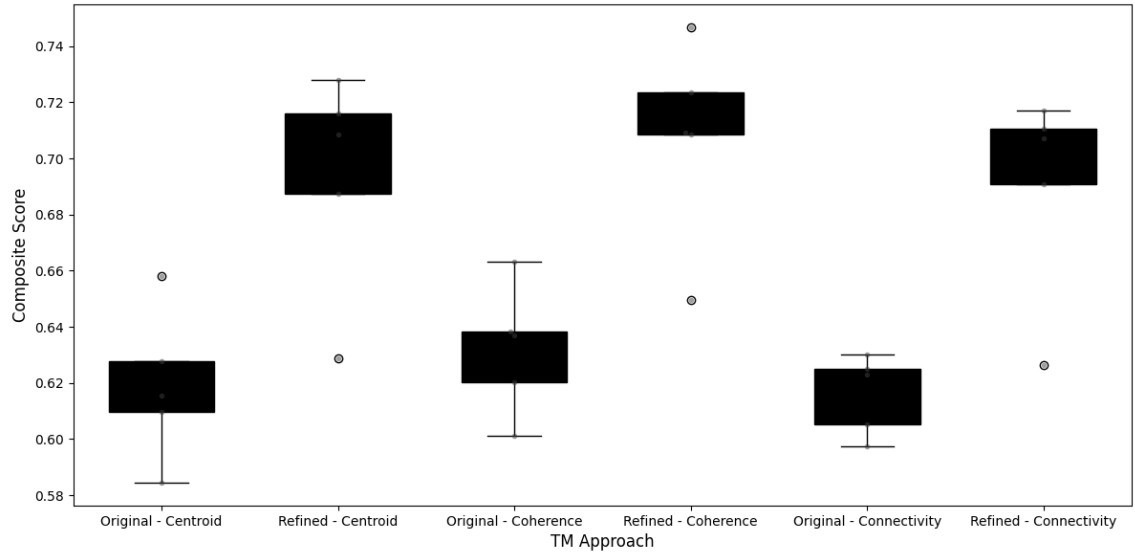

## B.1.6. French Dataset 3

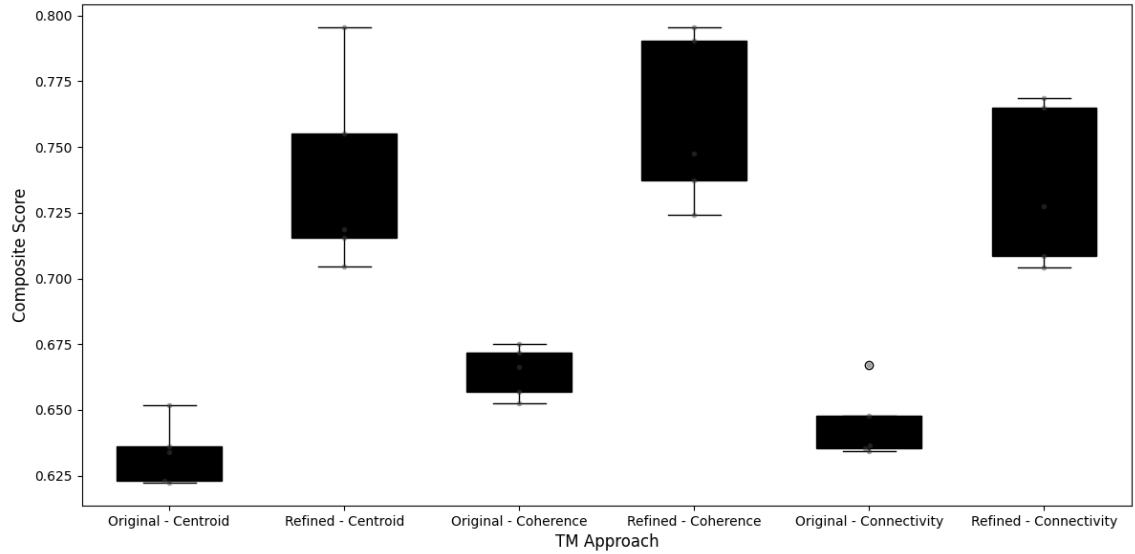

## B.2. Example of HAMLET's Topics – English Dataset 1 & French Dataset 1

| English Dataset 1 Topics |
|---|
| reception staff attitude |
| age-based assumptions |
| self-diagnosis |
| unhelpful website |
| premium rate telephone charges |
| poor emailing system |
| inappropriate dressing gown |
| poor service |
| nurses uncaring |
| abusive operator |
| admin staff performance |
| insufficient resources |
| interaction between staff and patients |
| unprofessional surgery |
| signage |
| reception staff rudeness |
| unprofessional staff |
| chasing documentation |
| outdated practices |
| non-helpful reception staff |
| unprofessional service |
| pain |
| poor reception service |
| muddled patient records |
| rushed consultations |
| lost test results |
| unhelpful reception staff |
| rubbish bins overflowing |
| parking issues |
| unresponsive phone service |
| two-week wait for appointment |
| carers not regularly sitting with patients |
| undermining family values |
| lights out timing |
| lack of high chairs for disabled people |
| nursing staff attitudes |
| respect for the patient |
| need for staff training on dysphasia |
| inconvenient opening hours |
| dirty waiting room |
| reduced parking capacity |
| new staff |
| gp registration |
| receptionist training |
| rude reception staff |
| staff were very unfriendly |
| staff scheduling |
| discharged in pyjamas |
| attitude of staff |
| difficulty reaching switchboard |

| | |
|---|---|
| reception staff attitude | |

| **French Dataset 1** | **English Translation** |
|---|---|
| situation politique | political situation |
| masques | masks |
| football | football |
| impact éducatif | educational impact |
| désinformation médicale | medical misinformation |
| événements en ligne | online events |
| politique belge | Belgian politics |
| foyer de contamination | contamination cluster |
| fête illégale | illegal party |
| anticorps | antibodies |
| tests salivaires | saliva tests |
| colère et pessimisme | anger and pessimism |
| variant brésilien | Brazilian variant |
| covid-19 dans les écoles | covid-19 in schools |
| recherche médicale | medical research |
| vente à emporter | takeaway sales |
| cancer | cancer |
| progrès médical | medical progress |
| festival | festival |
| respect des protocoles sanitaires | respect for health protocols |
| brésil | Brazil |
| spectacles virtuels | virtual shows |
| taux de positivité | positivity rate |
| vaccin pfizer | Pfizer vaccine |
| décès | deaths |
| levée des restrictions | lifting of restrictions |
| déconfinement | lockdown easing |
| violence | violence |
| covid-19 en martinique | covid-19 in Martinique |
| cyclisme | cycling |
| état de santé | health condition |
| impact sur l'équipe | impact on the team |
| covid-19 en île-de-france | covid-19 in √éle-de-France |
| événement sportif | sporting event |
| impact covid-19 sur le sport | covid-19 impact on sports |
| fête de thanksgiving | Thanksgiving celebration |
| enjeux climatiques | climate issues |
| restrictions de voyage | travel restrictions |
| états-unis | United States |
| certificat covid | covid certificate |
| préparation insuffisante | insufficient preparation |
| conditions de détention | detention conditions |
| contamination dans les bars | contamination in bars |
| équilibre épidémique | epidemic balance |
| reprise économique | economic recovery |
| biopolitique | biopolitics |
| événements annulés | canceled events |
| événements test | test events |
| épidémie au royaume-uni | epidemic in the United Kingdom |
| Situation épidémique | Epidemic situation |